%% file: ParetoGS-main.tex
\documentclass{article}

\usepackage{iclr2027_conference,times}
\usepackage{amsmath,amssymb,bm}
\usepackage{booktabs,multirow,graphicx,longtable}
\usepackage[table]{xcolor}
\usepackage{placeins}
\usepackage{float}
\definecolor{ParetoRow}{HTML}{EEF4FA}
\definecolor{ParetoAccent}{HTML}{25486B}
\definecolor{SceneTint}{HTML}{E8F0F8}
\definecolor{SceneRule}{HTML}{D0DAE4}
\newcommand{\SceneSeparator}{\arrayrulecolor{SceneRule}\specialrule{0.25pt}{0.65pt}{0.65pt}\arrayrulecolor{black}}
\usepackage{hyperref}
\usepackage{url}

\newcommand{\ParetoGSTitle}{ParetoGS: High-Quality Continuous-Budget 3D Gaussian Splatting with Sparse Adapted Anchors}
\title{\raggedright\hyphenpenalty=10000\ParetoGSTitle}
\author{Xiaobin Deng \quad Changyu Diao\thanks{Corresponding authors.} \quad Min Li \quad Duanqing Xu\footnotemark[1]\\{\normalfont Zhejiang University}}
\iclrfinalcopy
\hypersetup{hidelinks,pdftitle={\ParetoGSTitle},pdfauthor={Xiaobin Deng, Changyu Diao, Min Li, Duanqing Xu}}

\begin{document}
\maketitle
\fancyhead{}
\renewcommand{\headrulewidth}{0pt}
\input{sections/abstract}
\input{sections/01_introduction}
\input{sections/02_related_work}
\input{sections/03_method}
\input{sections/04_experiments}
\input{sections/05_conclusion}
\input{sections/statements}

\bibliography{ParetoGS-main}
\bibliographystyle{iclr2027_conference}
\clearpage
\appendix
\input{sections/appendix}

\end{document}

%% file: sections/abstract.tex
\begin{abstract}
Scalable 3D Gaussian Splatting allows one scene representation to serve different Gaussian-count budgets, yet its quality--budget curve may lag behind models optimized separately for each budget.
Reducing the number of Gaussians changes both the selected members and the reconstruction tasks carried by the survivors; parameters inherited from a high-budget model may fit a low-budget representation poorly.
We propose ParetoGS, which describes member changes and parameter evolution with a small set of adapted budget anchors and decodes intermediate budgets without separate optimization at every requested count.
ParetoGS alternates pruning and fixed-member adaptation to build nested anchors with persistent identities, and encodes their parameter changes relative to the full model in a compact residual atlas.
At decoding, it interpolates the states of shared members and gradually introduces the rendering contributions of new members, connecting sparse anchors into a continuous-budget representation.
Across 13 scenes, ParetoGS attains better average PSNR, SSIM, and LPIPS with fewer Gaussians than the compared prefix methods at 10\% of each method's own full model.
From a shared full model, it also outperforms FlexGS-imp at its trained 1\%, 5\%, and 10\% budgets and NanoGS-imp at the 1\% and 0.1\% budgets.
On Garden, its non-anchor quality approaches an empirical reference from independently optimized GNS models at higher retention ratios and exceeds it at some lower-budget points.
Across 13 scenes, encoding reduces mean additional state from 390.0 to 85.7 MB with an average PSNR change of 0.01 dB.
\par\noindent Project page: {\urlstyle{same}\url{https://xiaobin2001.github.io/ParetoGS-Web/}}.
\end{abstract}

%% file: sections/01_introduction.tex
\section{Introduction}
\label{sec:introduction}

\input{figures/method_overview}

The explicit primitives of 3D Gaussian Splatting (3DGS)~\citep{kerbl20233d} make Gaussian count a direct control on rendering cost.
Selecting a smaller subset from a full model, however, need not yield a high-quality representation at that count.
Pruning and re-optimizing for each budget gives strong individual models but requires repeated training and multiple assets.
Scalable methods instead use member orderings, hierarchies, or budget-conditioned transformations to serve many counts from one asset.
Their quality--budget curves can still lag behind models optimized separately at each count.
We ask whether one representation can decode any Gaussian-count budget while approaching this empirical per-budget quality reference.

Count reduction changes both membership and the reconstruction tasks of the survivors.
After other Gaussians are removed, retained members must adjust their geometry, opacity, and appearance to fit the reduced capacity.
Even a suitable subset can perform poorly with parameters inherited from the full model.
Its adapted state also provides a new starting point for the next pruning decision.
ParetoGS therefore contracts from previously adapted anchors and describes both member changes and parameter evolution.

Persistent identities retained through progressive contraction align the same Gaussian across neighboring budgets.
Budget evolution then has two parts: discrete member entries and exits, described by a nested order, and changes in the parameters of surviving members.
A few optimized budget anchors provide endpoints for the latter.
Aligned states at adjacent anchors define a piecewise parameter trajectory for estimating shared-member parameters at intermediate counts.

We propose ParetoGS (Figure~\ref{fig:pareto-overview}), which alternates structural contraction and fixed-member adaptation to build strictly nested anchors.
It preserves identities and deletion order, encodes each anchor's changes relative to the full model in a layered residual atlas, and decodes intermediate budgets by interpolating shared-member states and gradually activating entering members.
An approximately doubling schedule places anchors from 0.1\% of Full to the full count, yielding a single asset that serves every integer count in this range without further optimization.

Across 13 scenes, ParetoGS outperforms CLOD-3DGS and MGS at 10\% of each method's own full model while using fewer Gaussians.
From a shared full model, it improves mean quality over FlexGS-imp at its trained 1\%, 5\%, and 10\% budgets and over NanoGS-imp at 1\% and 0.1\%.
On Garden, non-anchor quality approaches independently optimized GNS results at higher retention and exceeds them at some lower budgets; qualitative results also show better low-budget structure.
Across 13 scenes, encoding reduces mean additional state from 390.0 to 85.7 MB.

Our contributions are:
\begin{itemize}
\item A representation of continuous-budget 3DGS as a joint trajectory of nested membership and budget-specific parameter states, organized by progressively adapted anchors.
\item ParetoGS, which connects those anchors through persistent identities, aligned residual encoding, and piecewise decoding to generate any intermediate count without separate optimization.
\item Thirteen-scene multi-budget comparisons, non-anchor evaluation, and ablations showing the value of adapted states and interval decoding, together with compact storage of the anchor family.
\end{itemize}

%% file: figures/method_overview.tex
\begin{figure}[t]
\centering
\includegraphics[width=\linewidth]{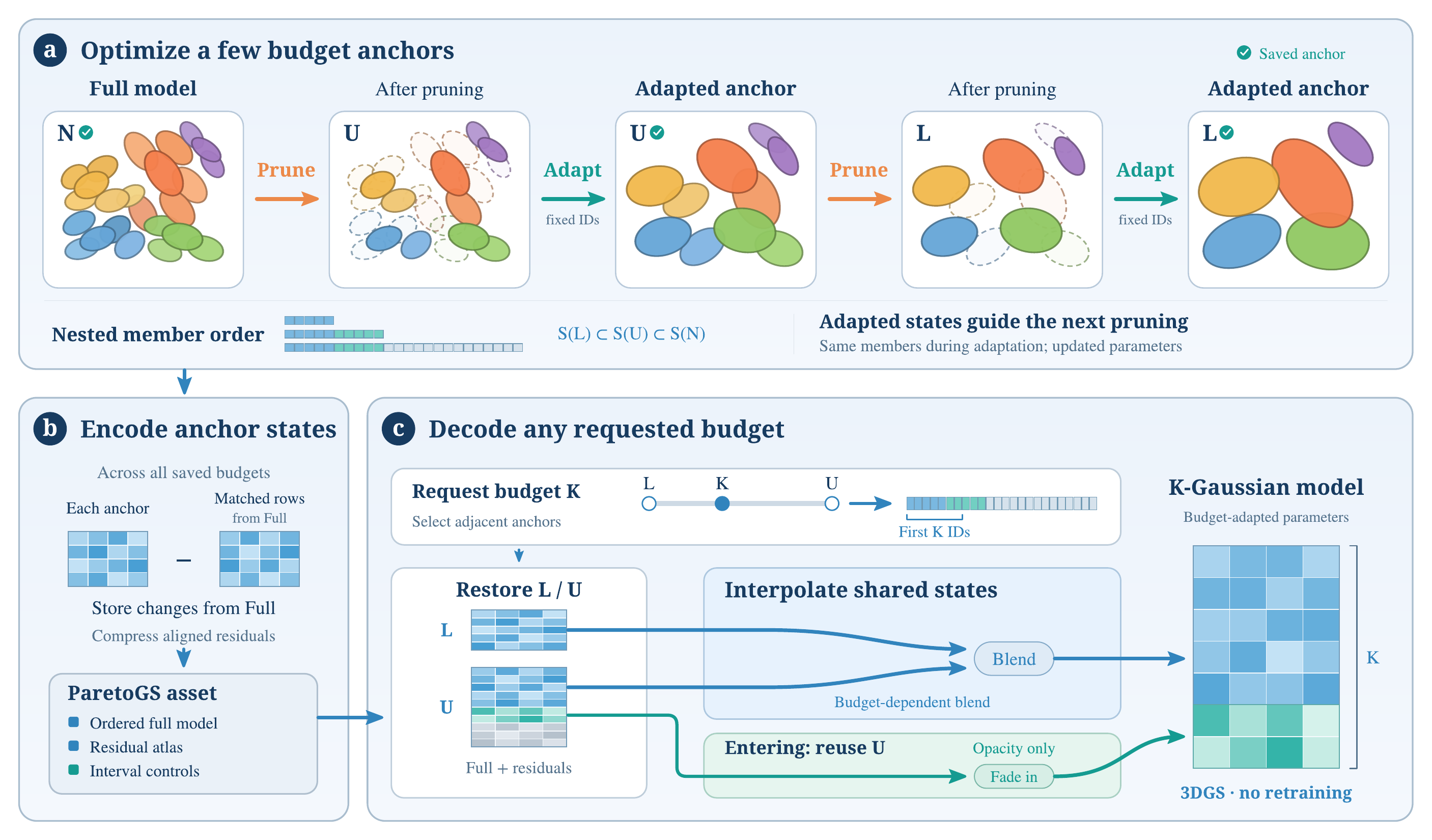}
\caption{\textbf{ParetoGS overview.}
\textbf{(a)} Alternating pruning and fixed-member adaptation builds sparse nested anchors; each adapted state seeds the next contraction.
\textbf{(b)} Persistent identities align anchor states, whose parameter changes relative to the full model are encoded in a compact residual atlas.
\textbf{(c)} For a requested count $K$ between adjacent anchors $L$ and $U$, decoding interpolates shared-member parameters and introduces $K-L$ entering members from $U$ with opacity gating, producing a standard $K$-Gaussian model.}
\label{fig:pareto-overview}
\end{figure}

%% file: sections/02_related_work.tex
\section{Related Work}
\label{sec:related_work}

\paragraph{3DGS representation and reconstruction.}
3DGS~\citep{kerbl20233d} combines explicit Gaussians with differentiable rasterization.
Subsequent work extends primitives and dynamic scenes~\citep{huang2024twodgs,wu2024fourdgs}, improves surface reconstruction~\citep{yu2024gof,chen2025pgsr}, visibility and filtering~\citep{radl2024stopthepop,yu2024mip}, rendering and training efficiency~\citep{hanson2025speedy,mallick2024taming}, and initialization and capacity allocation~\citep{kotovenko2026edgs,deng2025improved,kheradmand2024mcmc}.
Wild-GS, PPISP, and Deblurring 3D-GS address appearance changes, camera photometry, and blur in real captures~\citep{xu2024wildgs,deutsch2026ppisp,lee2024deblurring}.
We focus on organizing an already trained scene into an asset decodable at arbitrary Gaussian budgets.

\paragraph{Compact and multilevel representations.}
Pruning reduces membership, whereas compression encodes attributes more compactly.
HAC, SOGS, CompGS, and Compact-3DGS explore context coding, regular grids, vector quantization, and compact parameterization~\citep{chen2024hac,morgenstern2025sogs,navaneet2025compgs,lee2024compact}.
LightGaussian, PUP 3D-GS, and Trimming the Fat use contribution or sensitivity for pruning~\citep{fan2024lightgaussian,hanson2025pup,ali2024trimming}; Mini-Splatting and GHAP use sampling or mixture reduction~\citep{fang2024mini,wang2025ghap}.
MaskGaussian and GNS integrate member selection with adaptation~\citep{liu2025maskgaussian,deng2025gns}.
Scale and spatial hierarchies organize models at different resolutions or capacities~\citep{yan2024multiscale,ren2025octree,liu2025city,kerbl2024hierarchical,kulhanek2025lodge,seo2025flod,shen2025lodgs,cheng2026clodgs}.
H3DGS also interpolates optimized parent--child attributes~\citep{kerbl2024hierarchical}.
Contribution orderings and layered decompositions yield cumulative representations~\citep{zoomers2025progs,tsai2025l3gs,lavi2026frequency}, sometimes with progressive compression or attribute precision~\citep{disario2025gode,chen2026pcgs}.
ParetoGS uses GNS to build budget anchors and connects their aligned parameter states.

\paragraph{Arbitrary-budget decoding and adaptation.}
CLOD-3DGS fine-tunes random prefixes, MGS jointly trains a prefix and the full set on each view, and Differentiable Hierarchies learns a count-truncatable ordering~\citep{milef2025clod,guo2026mgs,pan2026hierarchies}.
These approaches reuse the same Gaussian parameters across prefix budgets.
RAVE reorders incremental members between neighboring bitrate levels to connect intermediate rates and counts~\citep{tran2026rave}.
LapisGS arranges content into cumulative resolution layers and interpolates readapted old-layer opacity while retaining old members' geometry and appearance~\citep{shi2025lapisgs}.
SCAR-GS progressively refines neural anchor features with residual vector quantization~\citep{revilla2026scar}; its residuals approximate features rather than record optimized states at different budgets.
NanoGS progressively merges local pairs to produce a compact model at a target count without image-supervised optimization~\citep{xiong2026nanogs}.
FlexGS learns both budget-conditioned selection and geometry transformation; its paper optimizes these modules after 15K pretraining~\citep{liu2025flexgs}.
ParetoGS instead stores a few explicitly optimized nested anchors, continues pruning from each adapted state, and connects the geometry, opacity, and appearance of corresponding members.
Experiments assess the resulting quality at equal own-Full fractions, from a shared Full, and at non-anchor budgets.

%% file: sections/03_method.tex
\section{Method}
\label{sec:method}

ParetoGS progressively contracts a trained full model into nested, adapted budget anchors.
Persistent identities align their members; a layered residual atlas stores anchor states relative to the full model.

\subsection{Budget-Conditioned Scene Representation}
\label{sec:budget-representation}

3DGS projects Gaussians carrying position, shape, opacity, and appearance into a view and composites them in depth order.
Their opacity and preceding transmittance determine the color at pixel $u$:
\begin{equation}
I(u)=\sum_i T_i(u)\widetilde a_i(u)c_i(u)
+T_{\mathrm{bg}}(u)c_{\mathrm{bg}},
\qquad
T_i(u)=\prod_{k\prec_u i}\bigl(1-\widetilde a_k(u)\bigr).
\end{equation}
Here $\widetilde a_i(u)$ combines activated opacity and projected coverage, $c_i(u)$ is view-dependent SH color, and $k\prec_u i$ denotes a Gaussian preceding $i$.
$T_i(u)$ is the transmittance before $i$; $T_{\mathrm{bg}}$ and $c_{\mathrm{bg}}$ describe the background.
Removal also changes later members' transmittance weights, so full-model parameters need not fit a contracted model.

Let the full model contain $N$ Gaussians, each with a persistent identity $z_i$ and parameter vector $\theta_i=(\mathbf x_i,\mathbf s_i,\mathbf r_i,o_i,\mathbf c_i)$ for position, log scale, raw rotation quaternion, opacity logit, and spherical-harmonic coefficients.
At budget $K$, let $\mathcal S(K)$ be the active identity set and $\theta_i(K)$ the parameters of member $i$.
The budget-conditioned model has exactly $K$ active members:
\begin{equation}
\mathcal G(K)=\left\{(z_i,\theta_i(K))\mid z_i\in\mathcal S(K)\right\},
\qquad |\mathcal S(K)|=K.
\end{equation}
We impose $\mathcal S(K_a)\subset\mathcal S(K_b)$ for $K_a<K_b$: larger budgets add members while allowing retained members' parameters to change.
States are connected in optimization coordinates; scales are exponentiated, opacity is passed through a sigmoid, and quaternions are normalized for rendering.
The asset decodes exactly $K$ Gaussians for every integer $K\in[K_{\min},N]$.

\subsection{Sparse Adapted Anchors and Member Correspondence}
\label{sec:budget-anchors}

We explicitly optimize a few budgets, adapting survivors at each anchor before starting the next contraction.

Let $M$ anchors have budgets $K_{\min}=K_1<K_2<\cdots<K_M=N$.
From Full, each stage contracts to the next target while optimizing parameters, then holds membership fixed and continues optimizing pixel $L_1$ and SSIM reconstruction losses.
The adapted state is a decoding endpoint and the next stage's starting point.
The chain deletes existing Gaussians without further densification; persistent IDs maintain correspondence as parameters change.

When contracting upper anchor $U$ to lower anchor $L$, we record deletions and reverse them to obtain the entry order $\pi_{L\rightarrow U}$ as budget grows.
For $L<K<U$, decoding retains every member of the lower anchor and admits the first $K-L$ entries:
\begin{equation}
\mathcal S(K)=\mathcal S(L)\cup\pi_{L\rightarrow U}[1:K-L],
\qquad L<K<U.
\end{equation}
Concatenating the smallest anchor with these entry sequences yields a global nested order, indexed so budget $K$ uses the first $K$ rows.

Our approximately doubling schedule starts at 0.1\% of the actual full count and ends at the full budget, giving 11 anchors for every scene.
The target counts are $K_j=\operatorname{round}(N\rho_j)$ for $\rho_j=0.001,0.002,\ldots,0.512,1$.
When building downward from Full, each stage inherits the previous anchor's Gaussian parameters, optimizer state, and persistent IDs.
It resets the iteration label to 14,999, uses GNS~\citep{deng2025gns} to reach $K_j$ during 15K--18K, and continues fixed-member adaptation through 25K.
GNS adjusts regularization in response to a prescribed linear opacity-decay trajectory at the budget boundary, advances removal within the selection window, and uses its budget safeguard to retain exactly $K_j$ members.
The Full model uses ImprovedGS~\citep{deng2025improved} densification; subsequent anchor construction follows GNS parameter optimization and pruning with densification disabled.

\subsection{Piecewise Budget Decoding}
\label{sec:budget-decoding}

The nested order fixes membership.
Shared members have aligned states at both neighboring anchors, whereas entering members have only an upper-anchor state.

Let $L<K<U$ denote lower and upper anchors and the requested count.
The first $L$ members connect their endpoint parameters with a weight that grows toward $U$.
For normalized progress $t\in(0,1)$,
\begin{equation}
\label{eq:pareto-interval}
t=\frac{K-L}{U-L},\qquad
\theta_i(K)=(1-t^p)\theta_i(L)+t^p\theta_i(U),
\quad i\leq L.
\end{equation}
$\theta_i(L)$ and $\theta_i(U)$ describe the same member.
The exponent $p>0$ maps budget progress onto their parameter-space segment: $p=1$ is linear, while $p<1$ and $p>1$ approach the upper state earlier and later.

Entering members take upper-anchor geometry and appearance; the shared interval gate $t^q$ controls their rendering contribution.
With $q>0$ and upper-anchor opacity $a_i(U)=\sigma(o_i(U))$,
\begin{equation}
a_i(K)=\operatorname{clip}\bigl(t^q a_i(U),\epsilon,1-\epsilon\bigr),
\qquad L<i\leq K.
\end{equation}
$\sigma$ is the sigmoid, and a small $\epsilon>0$ bounds opacity within $[\epsilon,1-\epsilon]$ before conversion back to a finite logit.
For example, decoding 150K between 100K and 200K anchors interpolates the first 100K states and admits 50K upper-anchor members with opacity scaled by $0.5^q$.

We calibrate $p,q$ per interval on training views while fixing anchors and member order.
At $t\in\{0.25,0.5,0.75\}$, candidate exponents minimize reconstruction loss plus penalties for loss regressions and slope deviations; Appendix~\ref{app:pareto-method} gives the objective and search grid.
Only two scalars are stored per interval.
Calibration uses FP16 residual anchor states; deployment restores compressed endpoints with the same exponents.
An anchor-budget request returns its endpoint directly.

\subsection{Layered Residual Atlas}
\label{sec:residual-atlas}

The layered residual atlas encodes all anchor changes by member layer and parameter field, with Full as a common reference; decoding a request restores only its neighboring endpoints.

Let $\Theta_{K_j}$ be the parameter matrix at anchor $K_j$, one Gaussian per row, and order the full matrix $\Theta_N$ by the global member sequence.
For an anchor, we store its changes relative to the corresponding rows of the full model:
\begin{equation}
\Theta_{K_j}=\Theta_N[1:K_j]+\Delta_j.
\end{equation}
The full anchor has zero residual and needs no additional state.

Members first present at anchor $K_s$ form layer $s$:
$\mathcal B_s=\mathcal S(K_s)\setminus\mathcal S(K_{s-1})$, with $K_0=0$ and $\mathcal S(K_0)=\varnothing$.
For each field $f$ and $s<M$, flattening and stacking residuals from $K_s$ through $K_{M-1}$ gives $D_s^f$, with anchors as rows and member components as columns.
Members present only at Full require no residual.

An orthogonal transform along the state dimension gives shared residual bases:
\begin{equation}
D_s^f=C_s^f B_s^f.
\end{equation}
Rows of $B_s^f$ are bases, and rows of $C_s^f$ are anchor coefficients.
We retain all state components, applying later compression to the bases.

Higher-order SH coefficients dominate storage.
For each residual basis component, we approximate its Gaussian-by-channel matrix along the channel axis, saving per-Gaussian coefficients and a shared channel basis that reconstructs every channel.
These coefficients and other fields' bases are block-quantized; anchor coefficients, channel bases, and quantization scales use half precision.
Appendix~\ref{app:pareto-method} gives the ranks and quantization rule.

The anchor schedule also bounds the amount of state to encode:
$\sum_{j<M}K_j\approx1.023N$, so the valid raw parameters of all non-full anchors approach one full parameter tensor before encoding.
The final asset comprises the ordered full model, residual atlas, interval exponents, and member-order metadata.
At runtime, an anchor residual $\widehat\Delta_j$ restored from the atlas is added to the corresponding reference rows:
$\widehat\Theta_{K_j}=\Theta_N[1:K_j]+\widehat\Delta_j$.
The hat denotes a state approximated by lossy encoding.
The resulting endpoints follow the rules of Section~\ref{sec:budget-decoding} to produce a standard $K$-Gaussian model for an existing rasterizer.
An anchor request incurs encoding error but no interval state-estimation error.
With the full model and required endpoints resident, parameter generation takes $O(K)$ work for fixed parameter dimension.

%% file: sections/04_experiments.tex
\section{Experiments}
\label{sec:experiments}

We evaluate cross-budget reconstruction quality, budget-specific parameter states, interval-decoding components, and additional storage to test whether sparse adapted anchors form a high-quality family of budgeted models.

\subsection{Experimental Setup}
\label{sec:pareto-setup}

\paragraph{Datasets and metrics.}
We report PSNR, SSIM~\citep{wang2004ssim}, and LPIPS~\citep{zhang2018lpips} on nine Mip-NeRF 360~\citep{barron2022mipnerf360} scenes, Train and Truck from Tanks and Temples~\citep{knapitsch2017tanks}, and Playroom and DrJohnson from Deep Blending~\citep{hedman2018deepblending}.
The outdoor and indoor Mip-NeRF 360 scenes use the $4\times$ and $2\times$ downsampled images provided by 3DGS~\citep{kerbl20233d}; the other datasets use its provided versions.
After sorting image filenames, every eighth view is held out for testing and the rest are used for training.
All methods use the same test views and image sizes, with the three metrics computed by the 3DGS evaluation implementation.
The main tables and component and encoding ablations weight the 13 scenes equally; ablations first average ten interval midpoints within each scene.

\paragraph{Implementation and comparison protocol.}
We train a full model for 30K iterations with ImprovedGS~\citep{deng2025improved}, then progressively contract and adapt it with GNS~\citep{deng2025gns}, retaining 11 budget anchors.
The anchors span 0.1\% to 100\% of each scene's full count; each interval uses ten training views to calibrate $p,q$.
Appendix~\ref{app:pareto-method} gives the member-order, state-encoding, and interval-calibration procedures.
All experiments use an RTX 4090 GPU with \texttt{data\_device} set to GPU.

We use the comparison methods' official source code in two protocols.
First, CLOD-3DGS~\citep{milef2025clod}, MGS~\citep{guo2026mgs}, and ParetoGS each use their own full model; we compare equal retention ratios and report actual Gaussian counts.
CLOD-3DGS trains down to 100K Gaussians.
Second, FlexGS-imp and NanoGS-imp receive the same ImprovedGS full anchor as ParetoGS, and we compare equal requested budgets.
FlexGS-imp runs the public FlexGS~\citep{liu2025flexgs} code's 35K-iteration budget-adaptation stage, replacing the output of its standard 3DGS coarse stage with the shared full model; this stage samples 1\%, 5\%, 10\%, and 15\%.
Table~\ref{tab:pareto-shared} also queries 0.1\% and 30\% outside that sampled range.
NanoGS-imp uses its official implementation to simplify the same full model offline~\citep{xiong2026nanogs}.
Table headers report 13-scene mean offline time: ParetoGS includes full-model and anchor training, encoding, calibration, and packaging, while FlexGS-imp includes the shared full-model construction and its 35K adaptation stage.
Appendix~\ref{app:pareto-scenes} provides per-scene quality, counts, and time.

\subsection{Reconstruction Quality and Offline Cost}

\input{tables/prefix_results}
\input{tables/shared_results}

\paragraph{Quality at equal retention ratios.}
Table~\ref{tab:pareto-prefix} shows the largest ParetoGS gains at low budgets.
At 10\%, 5\%, 1\%, and 0.1\%, all three mean quality metrics improve over CLOD-3DGS and MGS.
At 10\%, ParetoGS reaches 27.17 dB with about 162K Gaussians; CLOD-3DGS and MGS use about 303K and 492K for 25.59 and 25.74 dB.
Across those four low ratios, ParetoGS gains about 3.50 dB in PSNR over MGS on average.
CLOD-3DGS fine-tunes random prefixes after full pretraining, and MGS jointly optimizes the full set and a sampled prefix on each step.
Their mean offline costs are about 35 and 74 minutes, versus 10.3 minutes for ParetoGS's full model, anchor chain, encoding, calibration, and packaging.
At high budgets the differences narrow; MGS has slightly better mean LPIPS at Full and 30\%.

\paragraph{Quality from a shared full model.}
At the sampled 1\%, 5\%, and 10\% budgets in Table~\ref{tab:pareto-shared}, ParetoGS has better 13-scene mean PSNR, SSIM, and LPIPS than FlexGS-imp.
At 5\%, these are 26.48 versus 24.03 dB, 0.791 versus 0.712, and 0.315 versus 0.386.
FlexGS-imp spends about 81 minutes including construction of the shared full model and its 35K-step adaptation stage, whereas the ParetoGS asset takes about 10.3 minutes to construct.
The three trained-budget comparisons retain a clear quality gap.
NanoGS studies 10\%, 1\%, and 0.1\% as major compression budgets~\citep{xiong2026nanogs}.
At 1\%, ParetoGS achieves a mean PSNR of 24.40 dB, compared with 18.84 dB for NanoGS-imp.
At 0.1\%, PSNR rises from 16.37 to 21.15 dB, SSIM from 0.470 to 0.580, and LPIPS decreases from 0.637 to 0.539.
NanoGS-imp takes 48--56 seconds to convert the shared full model into each target model offline (Table~\ref{tab:pareto-shared}).
The default FlexGS-imp budget module has a fixed size of about 30 MB; Section~\ref{sec:pareto-storage} compares the additional ParetoGS states and total assets.

\input{figures/qualitative}

Figure~\ref{fig:pareto-qualitative} shows the same Garden test view across budgets.
At 30\% and 10\%, CLOD-3DGS and MGS reconstruct the table, flowerpots, and background well.
At 1\%, their main structures fragment, and at 0.1\% only scattered content remains.
FlexGS-imp retains the table and flowerpots at its trained 1\%, 5\%, and 10\% ratios, but much of the background becomes blurred.
NanoGS-imp retains more content around 1\% and 5\%, with dense jagged and fragmented artifacts.
At 0.1\%, CLOD-3DGS is below its 100K training-count floor and FlexGS-imp is below its smallest sampled training ratio; NanoGS-imp still preserves the scene outline.
ParetoGS produces clearer and more coherent table, flowerpots, and background at the decoded 30\%, 10\%, 5\%, and 1\% budgets.
Its smallest 0.1\% model comes from a progressively adapted anchor and retains the main objects and overall layout.
The visual results complement the quality comparisons in Table~\ref{tab:pareto-shared}.

\FloatBarrier
\subsection{Budget-Specific States and Continuous Decoding}

\input{figures/parameter_states}
\FloatBarrier

\paragraph{Why the same members need different states.}
The first row of Figure~\ref{fig:pareto-states} selects members at each budget while retaining their full-model parameters.
At 0.8\%, much of the table and background is missing, showing that member selection alone does not preserve reconstruction quality.
The second row fixes the same 24K Gaussians selected in the upper-left panel and changes only their budget-specific parameters.
Viewed from right to left, these members by themselves cover only a small part of the scene in their Full state; as budget decreases, they take on the table and background, eventually reconstructing the whole view at 0.8\%.
The large change in task for the same identities motivates adapted parameter states at each anchor and their connection between anchors.

\input{figures/continuous_budget}
\FloatBarrier

\paragraph{Quality between sparse anchors.}
Figure~\ref{fig:pareto-continuous} evaluates non-anchor budgets on Garden.
Independent GNS contracts the same Full directly to each target during 15K--18K and adapts fixed members through 25K; its measured points form an empirical per-budget reference.
ParetoGS uses actual decoded test-view renderings.
From 1\% to 50\% retention, its decoded SSIM and LPIPS closely track the independent GNS points.
Below about 1\%, the curves separate in favor of ParetoGS, with the largest gap at the smallest tested budgets.
The other 12 scenes in Appendix~\ref{app:pareto-curves} show a similar budget-dependent pattern.

\noindent\begin{minipage}{\linewidth}
\subsection{Decoder Components and Anchor Ablations}

We evaluate the arithmetic midpoint of each neighboring pair among the 11 anchors, giving ten budgets with identical member IDs across configurations.
Table~\ref{tab:pareto-decoding} compares full-model parameters for those members, a fixed lower-anchor shared state, no opacity gate for entering members, and linear progress $p=q=1$.
Without the gate, entering members use their upper-anchor opacity directly.
\end{minipage}\par

\input{tables/decoding_storage}

Fixing common members at the lower-anchor state reduces mean PSNR by 0.13 dB and SSIM by 0.0018, while increasing LPIPS by 0.0019.
Parameter changes matter even within a single anchor interval.
Without the entering-member gate, PSNR and SSIM fall while LPIPS stays similar.
Calibrated $p,q$ improve all three metrics over $p=q=1$.

Finally, assigning Full-model parameters to the same IDs lowers PSNR from 25.47 to 17.07 dB, SSIM from 0.7498 to 0.5459, and raises LPIPS from 0.3458 to 0.4718.
This quantifies the budget-specific state change illustrated in Figure~\ref{fig:pareto-states}; all 13 scenes show the same decline (Appendix Table~\ref{tab:pareto-decoding-scenes}).

\input{figures/anchor_density}

The decoded quality curves in Figure~\ref{fig:pareto-density} remain smooth when anchor spacing doubles or quadruples, with mean SSIM changing from 0.6569 to 0.6467 and 0.6180.
Directly adapted anchors exceed interpolation from sparse endpoints at 0.8\% and 3.2\%.
Even the two-endpoint interpolation at 5\% exceeds FlexGS-imp trained at that ratio (SSIM 0.6239, LPIPS 0.3894).
Intermediate anchors raise the connecting curve's average quality.

\subsection{State Encoding and Storage}
\label{sec:pareto-storage}

Across 13 scenes, the atlas reduces mean additional state from 390.0 MB (FP32) to 85.7 MB (atlas), or 78.0\%, with only a 0.01 dB PSNR drop, 0.0005 SSIM drop, and 0.0005 LPIPS increase (Table~\ref{tab:pareto-storage}).
FP16 residuals use 195.0 MB with the same quality at shown precision.
For Bicycle, Garden, and Stump, the atlas averages 158.7 MB (Appendix Table~\ref{tab:pareto-storage-scenes}); with their shared 744 MB Full, ParetoGS totals about 903 MB, versus about 774 MB for Full plus FlexGS-imp's 30 MB module.
This roughly 17\% asset increase accompanies better multi-budget quality and shorter offline construction.
With anchors resident on the GPU, ParetoGS parameter construction has a 2.77 ms P95 (Appendix~\ref{app:pareto-switch}).
A deployment capped at 51.2\% of Full can use that anchor as the reference and re-encode smaller states to reduce asset storage.

\FloatBarrier

%% file: tables/prefix_results.tex

\begin{table}[!ht]
\centering
\normalsize
\setlength{\abovecaptionskip}{\baselineskip}
\setlength{\belowcaptionskip}{\baselineskip}
\setlength{\tabcolsep}{2pt}
\setlength{\arrayrulewidth}{0.25pt}
\renewcommand{\arraystretch}{1.18}
\caption{Equal retention ratios from each method's own full model. Metrics and actual Gaussian counts are equally weighted 13-scene means; Num gives the actual count in thousands, and full-model sizes differ. Parentheses after method names give mean offline time in minutes; Ours denotes ParetoGS. Bold marks the best mean quality at each ratio.}
\label{tab:pareto-prefix}
\begin{tabular}{r|rrrr|rrrr|rrrr}
\toprule
\cellcolor[HTML]{E7EBF0} & \multicolumn{4}{c|}{\cellcolor[HTML]{DDEBF7}\textbf{CLOD-3DGS ($\sim$35 min)}} & \multicolumn{4}{c|}{\cellcolor[HTML]{E2EFDA}\textbf{MGS ($\sim$74 min)}} & \multicolumn{4}{c}{\cellcolor[HTML]{FCE4D6}\textbf{Ours ($\sim$10.3 min)}} \\
\cellcolor[HTML]{E7EBF0}\multirow{-2}{*}{\textbf{Ratio}} & \cellcolor[HTML]{DDEBF7}PSNR$\!\uparrow$ & \cellcolor[HTML]{DDEBF7}SSIM$\!\uparrow$ & \cellcolor[HTML]{DDEBF7}LPIPS$\!\downarrow$ & \cellcolor[HTML]{DDEBF7}Num & \cellcolor[HTML]{E2EFDA}PSNR$\!\uparrow$ & \cellcolor[HTML]{E2EFDA}SSIM$\!\uparrow$ & \cellcolor[HTML]{E2EFDA}LPIPS$\!\downarrow$ & \cellcolor[HTML]{E2EFDA}Num & \cellcolor[HTML]{FCE4D6}PSNR$\!\uparrow$ & \cellcolor[HTML]{FCE4D6}SSIM$\!\uparrow$ & \cellcolor[HTML]{FCE4D6}LPIPS$\!\downarrow$ & \cellcolor[HTML]{FCE4D6}Num \\
\midrule
\cellcolor[HTML]{F2F2F2}\textbf{100\%} & 27.54 & 0.847 & 0.197 & 3029 & 27.75 & 0.853 & \textbf{0.180} & 4923 & \textbf{27.89} & \textbf{0.854} & 0.185 & 1615 \\
\cellcolor[HTML]{F2F2F2}\textbf{30\%} & 27.24 & 0.834 & 0.222 & 909 & 27.32 & 0.835 & \textbf{0.206} & 1477 & \textbf{27.88} & \textbf{0.850} & 0.207 & 485 \\
\cellcolor[HTML]{F2F2F2}\textbf{10\%} & 25.59 & 0.779 & 0.294 & 303 & 25.74 & 0.757 & 0.301 & 492 & \textbf{27.17} & \textbf{0.821} & \textbf{0.269} & 162 \\
\cellcolor[HTML]{F2F2F2}\textbf{5\%} & 22.38 & 0.691 & 0.376 & 151 & 24.32 & 0.694 & 0.372 & 246 & \textbf{26.48} & \textbf{0.791} & \textbf{0.315} & 81 \\
\cellcolor[HTML]{F2F2F2}\textbf{1\%} & 12.86 & 0.383 & 0.581 & 30 & 20.35 & 0.565 & 0.507 & 49 & \textbf{24.40} & \textbf{0.697} & \textbf{0.422} & 16 \\
\cellcolor[HTML]{F2F2F2}\textbf{0.1\%} & 8.15 & 0.077 & 0.712 & 3 & 14.79 & 0.424 & 0.628 & 5 & \textbf{21.15} & \textbf{0.580} & \textbf{0.539} & 2 \\
\bottomrule
\end{tabular}
\end{table}

%% file: tables/shared_results.tex

\begin{table}[!ht]
\centering
\normalsize
\setlength{\abovecaptionskip}{\baselineskip}
\setlength{\belowcaptionskip}{\baselineskip}
\setlength{\tabcolsep}{2pt}
\setlength{\arrayrulewidth}{0.25pt}
\renewcommand{\arraystretch}{1.18}
\caption{Shared full initialization at equal requested budgets. Metrics and requested Gaussian counts are equally weighted 13-scene means; Num is in thousands, and Conv. is NanoGS-imp's offline conversion time from Full (seconds). Parentheses after method names give mean offline time in minutes; Ours denotes ParetoGS.}
\label{tab:pareto-shared}
\begin{tabular}{rr|rrr|rrrr|rrr}
\toprule
\cellcolor[HTML]{E7EBF0} & \cellcolor[HTML]{E7EBF0} & \multicolumn{3}{c|}{\cellcolor[HTML]{DDEBF7}\textbf{FlexGS-imp ($\sim$81 min)}} & \multicolumn{4}{c|}{\cellcolor[HTML]{E2EFDA}\textbf{NanoGS-imp}} & \multicolumn{3}{c}{\cellcolor[HTML]{FCE4D6}\textbf{Ours ($\sim$10.3 min)}} \\
\cellcolor[HTML]{E7EBF0}\multirow{-2}{*}{\textbf{Ratio}} & \cellcolor[HTML]{E7EBF0}\multirow{-2}{*}{\textbf{Num}} & \cellcolor[HTML]{DDEBF7}PSNR$\!\uparrow$ & \cellcolor[HTML]{DDEBF7}SSIM$\!\uparrow$ & \cellcolor[HTML]{DDEBF7}LPIPS$\!\downarrow$ & \cellcolor[HTML]{E2EFDA}PSNR$\!\uparrow$ & \cellcolor[HTML]{E2EFDA}SSIM$\!\uparrow$ & \cellcolor[HTML]{E2EFDA}LPIPS$\!\downarrow$ & \cellcolor[HTML]{E2EFDA}Conv. & \cellcolor[HTML]{FCE4D6}PSNR$\!\uparrow$ & \cellcolor[HTML]{FCE4D6}SSIM$\!\uparrow$ & \cellcolor[HTML]{FCE4D6}LPIPS$\!\downarrow$ \\
\midrule
\cellcolor[HTML]{F2F2F2}\textbf{30\%} & 485 & 21.98 & 0.705 & 0.355 & 23.86 & 0.708 & 0.329 & 48 & \textbf{27.88} & \textbf{0.850} & \textbf{0.207} \\
\cellcolor[HTML]{F2F2F2}\textbf{10\%} & 162 & 25.21 & 0.760 & 0.327 & 21.29 & 0.603 & 0.436 & 54 & \textbf{27.17} & \textbf{0.821} & \textbf{0.269} \\
\cellcolor[HTML]{F2F2F2}\textbf{5\%} & 81 & 24.03 & 0.712 & 0.386 & 20.34 & 0.564 & 0.486 & 55 & \textbf{26.48} & \textbf{0.791} & \textbf{0.315} \\
\cellcolor[HTML]{F2F2F2}\textbf{1\%} & 16 & 21.75 & 0.621 & 0.488 & 18.84 & 0.511 & 0.579 & 56 & \textbf{24.40} & \textbf{0.697} & \textbf{0.422} \\
\cellcolor[HTML]{F2F2F2}\textbf{0.1\%} & 2 & 13.54 & 0.392 & 0.641 & 16.37 & 0.470 & 0.637 & 56 & \textbf{21.15} & \textbf{0.580} & \textbf{0.539} \\
\bottomrule
\end{tabular}
\end{table}

%% file: figures/qualitative.tex
\begin{figure}[!t]
\centering
\includegraphics[width=\linewidth]{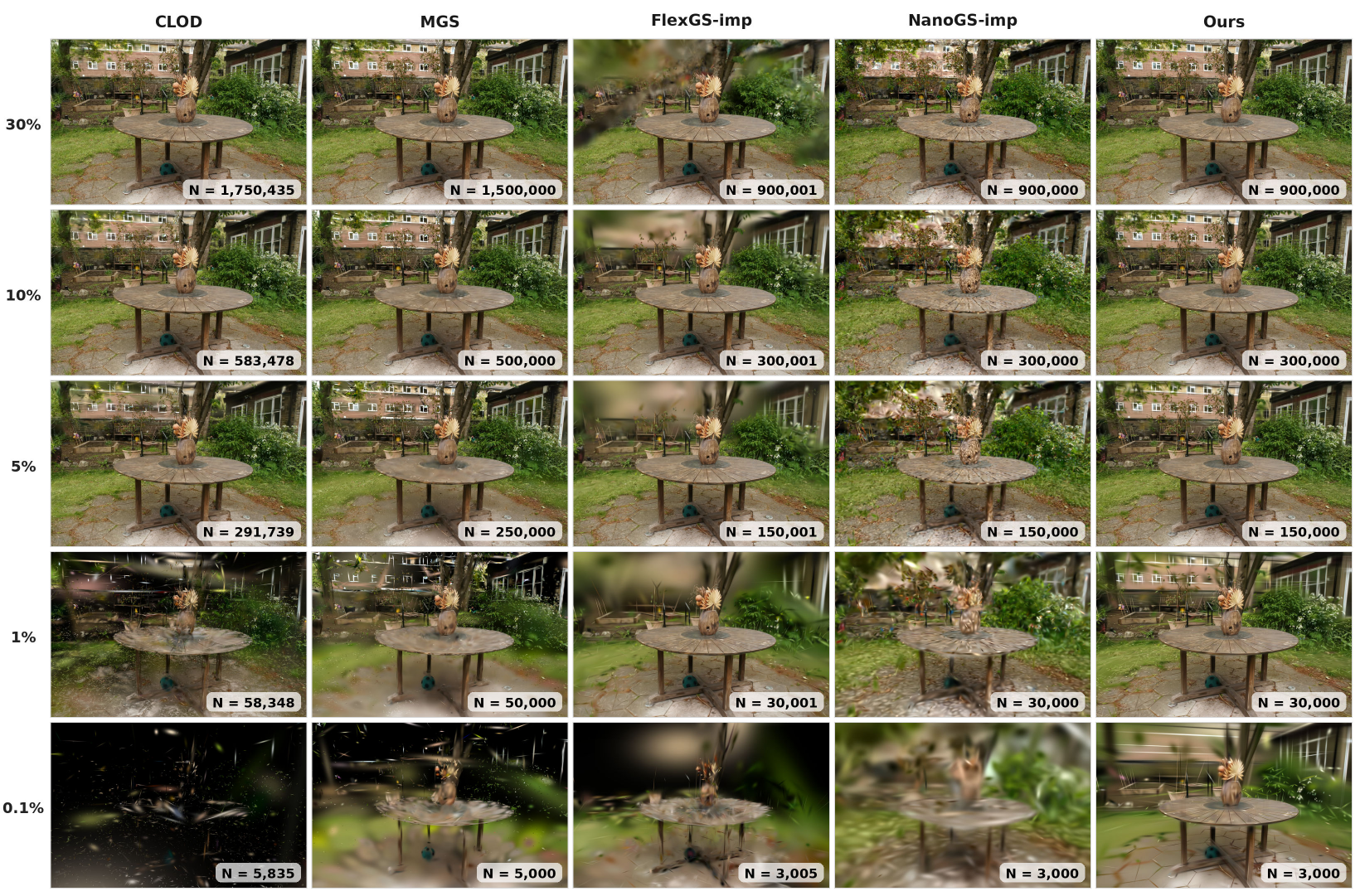}
\caption{Cross-budget reconstruction of the same Garden test view.
The five columns show CLOD-3DGS, MGS, FlexGS-imp, NanoGS-imp, and ParetoGS; the rows show 30\%, 10\%, 5\%, 1\%, and 0.1\%.
The first two columns use a fraction of each method's own full model; the last three use a shared full model.
Each panel labels its actual Gaussian count.}
\label{fig:pareto-qualitative}
\end{figure}

%% file: figures/parameter_states.tex
\begin{figure}[H]
\centering
\includegraphics[width=\linewidth]{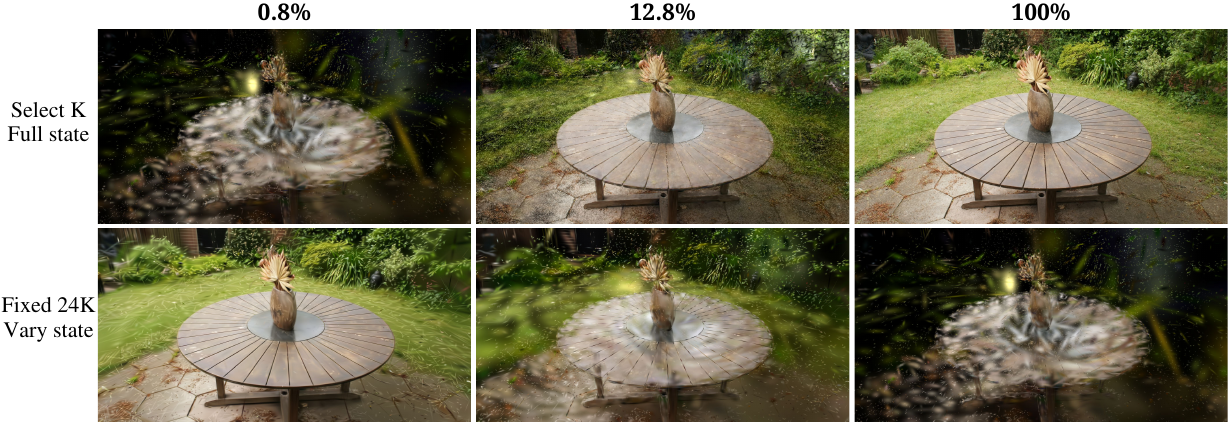}
\caption{Parameter states of the same members across budgets.
The Garden columns correspond to 0.8\%, 12.8\%, and 100\% of Full.
The first row selects 24K, 384K, and 3M members, respectively, while retaining Full parameters.
The second row fixes the same 24K members from the upper-left panel; column headings now indicate the budget from which their parameters were taken.
All panels use the same view and crop.}
\label{fig:pareto-states}
\end{figure}

%% file: figures/continuous_budget.tex
\begin{figure}[!ht]
\centering
\includegraphics[width=\linewidth]{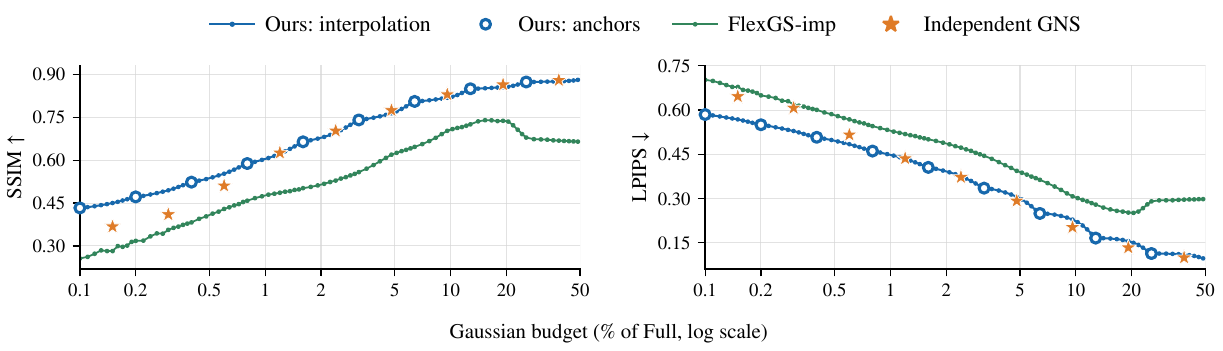}
\caption{Continuous-budget quality on Garden.
The full model has 3M Gaussians. SSIM and LPIPS are averaged over 24 test views against retention percentage on a log axis.
Blue open circles are ParetoGS anchors and small blue dots are decoded intervals; green denotes FlexGS-imp, and orange stars are independently optimized GNS references.
Line segments connect measured points over 0.1\%--50\%.}
\label{fig:pareto-continuous}
\end{figure}

%% file: tables/decoding_storage.tex

\begin{table}[!ht]
\centering
\normalsize
\setlength{\abovecaptionskip}{\baselineskip}
\setlength{\belowcaptionskip}{\baselineskip}
\setlength{\tabcolsep}{2pt}
\renewcommand{\arraystretch}{1.15}
\begin{minipage}[t]{.485\linewidth}
\vspace{0pt}
\begin{minipage}[t][4\baselineskip][t]{\linewidth}
\vspace{0pt}
\caption{Decoder components: mean quality at ten interval midpoints across 13 scenes. The Full-parameter row uses the same member IDs.}
\label{tab:pareto-decoding}
\end{minipage}\par
\centering
\begin{tabular}{lrrr}
\toprule
Variant & PSNR$\uparrow$ & SSIM$\uparrow$ & LPIPS$\downarrow$ \\
\midrule
\rowcolor{ParetoRow}
Full method & \textbf{25.47} & \textbf{0.7498} & 0.3458 \\
Fixed shared state & 25.35 & 0.7480 & 0.3477 \\
No opacity gate & 25.26 & 0.7478 & \textbf{0.3456} \\
Linear progress & 25.33 & 0.7459 & 0.3491 \\
Full parameters & 17.07 & 0.5459 & 0.4718 \\
\bottomrule
\end{tabular}
\end{minipage}\hfill
\begin{minipage}[t]{.485\linewidth}
\vspace{0pt}
\begin{minipage}[t][4\baselineskip][t]{\linewidth}
\vspace{0pt}
\caption{State encoding: quality at the same midpoints and mean additional storage across 13 scenes.}
\label{tab:pareto-storage}
\end{minipage}\par
\centering
\begin{tabular}{lrrrr}
\toprule
Encoding & PSNR$\!\uparrow$ & SSIM$\!\uparrow$ & LPIPS$\!\downarrow$ & MB$\!\downarrow$ \\
\midrule
FP32 & 25.49 & 0.7503 & 0.3452 & 390.0 \\
FP16 & 25.49 & 0.7503 & 0.3452 & 195.0 \\
\rowcolor{ParetoRow}
Atlas & 25.47 & 0.7498 & 0.3458 & 85.7 \\
\bottomrule
\end{tabular}
\end{minipage}
\end{table}

%% file: figures/anchor_density.tex
\begin{figure}[!ht]
\centering
\includegraphics[width=\linewidth]{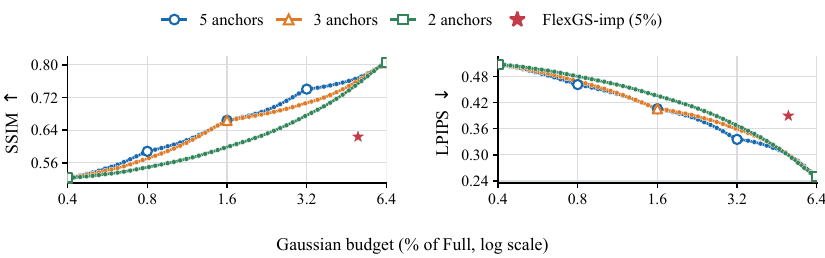}
\caption{Effect of intermediate anchors on connection quality.
From the same Garden state library and common 0.4\% and 6.4\% endpoints, we retain five, three, or two anchors.
Open circles, triangles, and squares mark their anchors; small solid points show real evaluations at 50 shared non-anchor budgets.
The red star is FlexGS-imp at the 5\% Garden budget from the shared full model.
Each configuration calibrates its own interval exponents, and metrics average all 24 test views.}
\label{fig:pareto-density}
\end{figure}

%% file: sections/05_conclusion.tex
\section{Conclusion}

ParetoGS describes both Gaussian membership and budget-specific parameter states with sparse adapted anchors.
Nested member sets and persistent identities establish a cross-budget member order, while shared-state interpolation and an entering-member gate connect discrete anchors into a family of models decodable at requested counts.
A residual atlas compresses the additional adapted states.

Across 13 scenes, ParetoGS achieves better mean quality than the compared prefix methods with fewer Gaussians at 10\% retention.
From a shared full model, it also improves over FlexGS-imp at its sampled 1\%, 5\%, and 10\% budgets and over NanoGS-imp at 1\% and 0.1\%.
Continuous-budget curves and component ablations across 13 scenes further demonstrate the value of adapted parameters and interval decoding.
The compressed atlas reduces mean additional state from 390.0 to 85.7 MB with little average quality change.
Organizing budget-specific states into one asset thus provides higher-quality multi-budget decoding from a single scene representation.
In a three-scene asset comparison, ParetoGS uses about 17\% more total storage than FlexGS-imp; its mean offline construction time across 13 scenes is shorter than those of CLOD-3DGS, MGS, and FlexGS-imp.

%% file: sections/statements.tex
\subsection*{AI Use Statement}

Generative AI tools assisted with code writing, figure and table preparation, manuscript revision and English translation, and provided feedback on the method, ablation design, and interpretation of results.
The authors reviewed all AI-assisted work, checked the algorithm, experimental data, and manuscript claims, and take responsibility for the final content.

\subsection*{Ethics Statement}

This work studies 3D reconstruction on established benchmark datasets. We are not aware of ethical issues specific to the proposed method or its evaluation.

\subsection*{Reproducibility Statement}

Section~\ref{sec:experiments} specifies the datasets and comparison protocol.
Appendix~\ref{app:pareto-method} provides the ordering, encoding, calibration, and decoding procedures; Appendix~\ref{app:pareto-scenes} reports per-scene results.
Appendix~\ref{app:pareto-switch} describes the budget-switch timing protocol and results.

%% file: sections/appendix.tex
\section{State Encoding, Calibration, and Budget Decoding}
\label{app:pareto-method}

This section specifies the member ordering, residual encoding, interval calibration, and decoding rules used by the main method.
Given constructed nested adapted anchors and the full model, the runtime procedure returns parameters for exactly $K$ Gaussians at any valid integer budget.

\subsection{Member Ordering and Residual Atlas}

When contracting an upper anchor $U$ to a lower anchor $L$, we record deleted IDs.
Reversing each deletion trace yields the order in which members enter as budget increases.
Starting with the smallest anchor and concatenating these entry sequences gives the global nested ordering used in the paper; full-model parameters are reordered accordingly.
Anchors are aligned to this ordering by ID, and their residuals relative to Full are computed in raw parameter coordinates.
The six fields are position, log scale, quaternion, opacity logit, DC color, and higher-order SH coefficients, giving 59 scalar components per Gaussian at SH degree three.

Non-full anchor residuals are converted to FP16 before stacking by member layer and parameter field.
Each layer undergoes an orthogonal decomposition along the anchor-state dimension.
The maximum supplied state-axis rank is ten; all $11-s$ available components of layer $s$ are retained, with no truncation along that dimension.
The 45 higher-order SH channels use a rank-36 channel subspace, while decoding still outputs all 45 channels.
For non-SH state bases and the per-Gaussian SH-channel coefficients, we apply symmetric INT8 quantization independently to each channel within blocks of 64 Gaussian rows:
\begin{equation}
s_{bc}=\frac{\max_{i\in b}|x_{ic}|}{127},\qquad
q_{ic}=\operatorname{clip}\!\left(\operatorname{round}\!\left(\frac{x_{ic}}{s_{bc}}\right),-127,127\right),\qquad
\widehat x_{ic}=q_{ic}\widehat s_{bc}.
\end{equation}
An all-zero block uses scale one.
Quantization scales, anchor combination coefficients, and higher-order SH channel bases use FP16.
Atlas construction consists of decomposition, quantization, and storage; the compressed states receive no further image optimization.

\subsection{Interval Calibration and Budget Decoding}

For each neighboring anchor pair, we enumerate the shared-state exponent $p$ and entering-member gate exponent $q$ over
\begin{equation}
\mathcal P=\{0.25,0.5,0.75,1,1.5,2,3,4,6\},\qquad
\mathcal Q=\{0.25,0.5,0.75,1,1.25,1.5,2,3\}.
\end{equation}
At rounded budgets corresponding to $t\in\{0.25,0.5,0.75\}$, we compare all 72 pairs on the same set of up to ten training views.
The views are midpoint selections from equally partitioned intervals of the training-camera list.
Calibration uses FP16 anchor states, does not update Gaussian parameters, and uses no test views.

Let $\ell_{v,h}$ be the reconstruction loss for view $v$ at the lower endpoint, three intermediate budgets, and upper endpoint, indexed by $h=0,\ldots,4$.
Let $t_h$ be the actual normalized progress of each rounded count.
For $V$ selected views, define $Q_+$ as the mean of the largest $\lceil V/4\rceil$ values in a set:
\begin{equation}
\begin{aligned}
d_{v,h}&=\frac{\ell_{v,h}-\ell_{v,h+1}}{t_{h+1}-t_h},\\
m_v&=\frac14\sum_{h=0}^{3}\max(-d_{v,h},0),\qquad
s_v=\frac14\sum_{h=0}^{3}|d_{v,h}-(\ell_{v,0}-\ell_{v,4})|,\\
J(p,q)&=\frac1{3V}\sum_{v=1}^{V}\sum_{h=1}^{3}\ell_{v,h}\\
&\qquad+Q_+(\{m_v\})+0.25Q_+(\{s_v\}).
\end{aligned}
\end{equation}
The image loss is $\ell=0.8L_1+0.2(1-\mathrm{SSIM})$.
Each interval stores the minimizing $p,q$; an exact tie favors larger $p$ and then larger $q$.
No per-Gaussian progress correction is stored.
Deployment restores endpoints from the compressed atlas while retaining the calibrated exponents.

An exact anchor request restores that anchor, and $K=N$ uses Full.
For other counts, shared members follow Equation~\eqref{eq:pareto-interval} in raw parameter coordinates.
Entering members use their upper-anchor state, multiply their activated opacity by $t^q$, clip it to $[10^{-6},1-10^{-6}]$, and convert it back to a logit.
The decoder outputs standard $K$-Gaussian parameters.
The deployed asset comprises the reordered Full, compressed atlas, interval exponents, and member-order metadata; by default, the two most recently used endpoints are cached.

\section{Per-Scene Experimental Results}
\label{app:pareto-scenes}

The following tables expand Tables~\ref{tab:pareto-prefix} and~\ref{tab:pareto-shared} into per-scene quality, Gaussian counts, and training, encoding, calibration, packaging, and full-construction time.
\input{tables/scene_results}

\subsection{Per-Scene Decoder Components}
Table~\ref{tab:pareto-decoding-scenes} expands Table~\ref{tab:pareto-decoding} over all 13 scenes.
\input{tables/decoding_scene}

\subsection{Per-Scene State Encoding}
Table~\ref{tab:pareto-storage-scenes} expands Table~\ref{tab:pareto-storage} into per-scene quality and additional state storage.
\input{tables/storage_scene}

\section{Budget-Switch Parameter Construction Time}
\label{app:pareto-switch}

We measure the time required to construct Gaussian parameters for a requested budget after the model has been loaded.
The four methods use the same 13 scenes and an RTX 4090 GPU, running serially in separate processes.
Before timing, CLOD-3DGS and MGS have their Full parameters and prefix orders resident on the GPU; FlexGS-imp has its Full parameters and selector; ParetoGS has all anchor fields predecoded and resident.
At a switch, CLOD-3DGS and MGS construct rendering parameters for the requested prefix, FlexGS-imp selects members with its native threshold and applies deformation, and ParetoGS interpolates endpoint states and gates entering members.
A Full request to ParetoGS reuses the resident model.

We synchronize CUDA and read host wall time $t_0$ before a request, then synchronize after target parameters are constructed and read $t_1$.
The reported time is $T_{\mathrm{switch}}=t_1-t_0$.
After each timed request, we render the first test view of that scene before the next request.
The trajectory covers 16 ratios from 0.1\% through 100\%, including forward, reverse, and cross-interval changes.
After three repetitions and removing requests with no change in ratio, each method contributes 108 switches per scene and 1,404 across 13 scenes.

\begin{table}[!ht]
\centering
\caption{Synchronized budget-switch parameter preparation time across 13 scenes (milliseconds), pooling 1,404 switches per method.}
\label{tab:pareto-switch-latency}
\begin{tabular}{lrrr}
\toprule
Method & Median & P95 & Maximum \\
\midrule
CLOD-3DGS & \textbf{0.12} & 3.01 & 7.46 \\
MGS & 0.41 & 7.79 & 8.10 \\
FlexGS-imp & 18.66 & 46.91 & 98.87 \\
ParetoGS & 0.27 & \textbf{2.77} & \textbf{5.65} \\
\bottomrule
\end{tabular}
\end{table}

ParetoGS has a median of 0.27 ms, P95 of 2.77 ms, and maximum of 5.65 ms in Table~\ref{tab:pareto-switch-latency}, showing low computational cost for endpoint interpolation and parameter construction.
Its one-time predecoding of all anchors averages 0.39 s, and the decoded endpoint fields occupy 390.0 MB of GPU storage on average.
An application can instead keep decoded endpoints in host memory or on disk and load them as needed according to its GPU memory budget; the table measures interpolation and parameter construction with endpoints already resident on the GPU.

\input{sections/appendix_visual}

\input{sections/appendix_curves}

%% file: tables/scene_results.tex

\begingroup
\setlength{\tabcolsep}{3pt}
\renewcommand{\arraystretch}{1.08}
\begin{longtable}{@{}llrrrrrr@{}}
\caption{Per-scene PSNR (dB) at retention ratios from each method's own Full.}\label{tab:pareto-scenes-prefix-psnr} \\
\toprule
Scene & Method & 100\% & 30\% & 10\% & 5\% & 1\% & 0.1\% \\
\midrule
\endfirsthead
\toprule
Scene & Method & 100\% & 30\% & 10\% & 5\% & 1\% & 0.1\% \\
\midrule
\endhead
\rowcolor{SceneTint}
Bicycle & CLOD-3DGS & 25.54 & 25.17 & 23.83 & 22.55 & 16.33 & 10.20 \\*
\rowcolor{SceneTint}
Bicycle & MGS & 25.50 & 24.73 & 22.35 & 20.88 & 18.76 & 14.28 \\*
\rowcolor{SceneTint}
Bicycle & ParetoGS & 25.87 & 25.81 & 25.18 & 24.63 & 23.36 & 21.62 \\
\SceneSeparator
\rowcolor{white}
Flowers & CLOD-3DGS & 21.87 & 21.47 & 20.16 & 18.96 & 12.23 & 7.73 \\*
\rowcolor{white}
Flowers & MGS & 21.89 & 21.11 & 18.73 & 17.65 & 15.40 & 11.73 \\*
\rowcolor{white}
Flowers & ParetoGS & 21.79 & 21.84 & 21.23 & 20.66 & 19.25 & 17.72 \\
\SceneSeparator
\rowcolor{SceneTint}
Garden & CLOD-3DGS & 27.93 & 27.54 & 25.83 & 23.93 & 15.28 & 8.16 \\*
\rowcolor{SceneTint}
Garden & MGS & 27.95 & 27.30 & 24.85 & 22.68 & 18.48 & 13.99 \\*
\rowcolor{SceneTint}
Garden & ParetoGS & 28.03 & 27.93 & 26.65 & 25.70 & 23.64 & 21.52 \\
\SceneSeparator
\rowcolor{white}
Stump & CLOD-3DGS & 27.16 & 26.95 & 25.77 & 24.37 & 16.65 & 12.08 \\*
\rowcolor{white}
Stump & MGS & 27.12 & 26.62 & 24.45 & 22.33 & 20.05 & 16.77 \\*
\rowcolor{white}
Stump & ParetoGS & 27.24 & 27.31 & 26.90 & 26.34 & 24.47 & 21.98 \\
\SceneSeparator
\rowcolor{SceneTint}
Treehill & CLOD-3DGS & 23.04 & 22.88 & 22.31 & 21.51 & 12.87 & 8.65 \\*
\rowcolor{SceneTint}
Treehill & MGS & 22.88 & 22.14 & 20.29 & 19.54 & 17.44 & 12.69 \\*
\rowcolor{SceneTint}
Treehill & ParetoGS & 22.81 & 23.16 & 22.97 & 22.75 & 22.02 & 20.80 \\
\SceneSeparator
\rowcolor{white}
Room & CLOD-3DGS & 32.36 & 32.03 & 30.22 & 24.20 & 11.96 & 8.87 \\*
\rowcolor{white}
Room & MGS & 32.05 & 31.80 & 30.72 & 29.56 & 24.19 & 16.05 \\*
\rowcolor{white}
Room & ParetoGS & 32.60 & 32.55 & 31.84 & 31.13 & 28.61 & 23.36 \\
\SceneSeparator
\rowcolor{SceneTint}
Counter & CLOD-3DGS & 29.47 & 29.09 & 26.95 & 20.86 & 11.98 & 8.91 \\*
\rowcolor{SceneTint}
Counter & MGS & 29.84 & 29.49 & 28.21 & 26.88 & 21.95 & 15.69 \\*
\rowcolor{SceneTint}
Counter & ParetoGS & 29.67 & 29.66 & 28.86 & 28.03 & 25.53 & 21.37 \\
\SceneSeparator
\rowcolor{white}
Kitchen & CLOD-3DGS & 32.06 & 31.57 & 29.13 & 23.95 & 11.91 & 6.20 \\*
\rowcolor{white}
Kitchen & MGS & 32.33 & 32.00 & 30.46 & 28.70 & 22.80 & 16.11 \\*
\rowcolor{white}
Kitchen & ParetoGS & 32.52 & 32.15 & 30.28 & 28.68 & 25.17 & 21.17 \\
\SceneSeparator
\rowcolor{SceneTint}
Bonsai & CLOD-3DGS & 32.60 & 32.13 & 28.42 & 20.46 & 12.09 & 9.54 \\*
\rowcolor{SceneTint}
Bonsai & MGS & 32.98 & 32.55 & 30.70 & 28.65 & 23.34 & 17.09 \\*
\rowcolor{SceneTint}
Bonsai & ParetoGS & 32.87 & 32.89 & 31.28 & 29.81 & 26.30 & 22.27 \\
\SceneSeparator
\rowcolor{white}
Train & CLOD-3DGS & 22.33 & 21.77 & 19.21 & 14.80 & 8.52 & 5.66 \\*
\rowcolor{white}
Train & MGS & 23.01 & 22.48 & 21.17 & 19.86 & 16.86 & 12.21 \\*
\rowcolor{white}
Train & ParetoGS & 22.36 & 22.48 & 22.21 & 21.86 & 20.41 & 17.48 \\
\SceneSeparator
\rowcolor{SceneTint}
Truck & CLOD-3DGS & 26.46 & 26.20 & 24.80 & 22.46 & 11.06 & 5.61 \\*
\rowcolor{SceneTint}
Truck & MGS & 26.40 & 26.15 & 24.82 & 23.13 & 19.03 & 13.84 \\*
\rowcolor{SceneTint}
Truck & ParetoGS & 26.44 & 26.52 & 25.98 & 25.32 & 23.12 & 19.47 \\
\SceneSeparator
\rowcolor{white}
Playroom & CLOD-3DGS & 28.52 & 28.87 & 28.48 & 26.99 & 11.28 & 5.45 \\*
\rowcolor{white}
Playroom & MGS & 29.66 & 29.66 & 29.49 & 28.92 & 23.09 & 14.96 \\*
\rowcolor{white}
Playroom & ParetoGS & 30.68 & 30.32 & 30.26 & 30.00 & 27.69 & 22.41 \\
\SceneSeparator
\rowcolor{SceneTint}
DrJohnson & CLOD-3DGS & 28.61 & 28.52 & 27.61 & 25.93 & 15.07 & 8.89 \\*
\rowcolor{SceneTint}
DrJohnson & MGS & 29.15 & 29.09 & 28.40 & 27.42 & 23.18 & 16.90 \\*
\rowcolor{SceneTint}
DrJohnson & ParetoGS & 29.73 & 29.81 & 29.61 & 29.30 & 27.68 & 23.75 \\
\bottomrule
\end{longtable}
\endgroup

\begingroup
\setlength{\tabcolsep}{3pt}
\renewcommand{\arraystretch}{1.08}
\begin{longtable}{@{}llrrrrrr@{}}
\caption{Per-scene SSIM at retention ratios from each method's own Full.}\label{tab:pareto-scenes-prefix-ssim} \\
\toprule
Scene & Method & 100\% & 30\% & 10\% & 5\% & 1\% & 0.1\% \\
\midrule
\endfirsthead
\toprule
Scene & Method & 100\% & 30\% & 10\% & 5\% & 1\% & 0.1\% \\
\midrule
\endhead
\rowcolor{SceneTint}
Bicycle & CLOD-3DGS & 0.7926 & 0.7680 & 0.6797 & 0.5945 & 0.3539 & 0.0728 \\*
\rowcolor{SceneTint}
Bicycle & MGS & 0.7939 & 0.7492 & 0.5655 & 0.4569 & 0.3460 & 0.2245 \\*
\rowcolor{SceneTint}
Bicycle & ParetoGS & 0.8050 & 0.7988 & 0.7531 & 0.7072 & 0.5728 & 0.4388 \\
\SceneSeparator
\rowcolor{white}
Flowers & CLOD-3DGS & 0.6381 & 0.6047 & 0.5146 & 0.4360 & 0.2184 & 0.0395 \\*
\rowcolor{white}
Flowers & MGS & 0.6451 & 0.6022 & 0.4427 & 0.3645 & 0.2545 & 0.1596 \\*
\rowcolor{white}
Flowers & ParetoGS & 0.6400 & 0.6272 & 0.5684 & 0.5165 & 0.3907 & 0.2859 \\
\SceneSeparator
\rowcolor{SceneTint}
Garden & CLOD-3DGS & 0.8777 & 0.8629 & 0.7947 & 0.7062 & 0.3760 & 0.0964 \\*
\rowcolor{SceneTint}
Garden & MGS & 0.8789 & 0.8562 & 0.7325 & 0.5887 & 0.3796 & 0.2574 \\*
\rowcolor{SceneTint}
Garden & ParetoGS & 0.8822 & 0.8739 & 0.8207 & 0.7683 & 0.6039 & 0.4328 \\
\SceneSeparator
\rowcolor{white}
Stump & CLOD-3DGS & 0.7991 & 0.7873 & 0.7250 & 0.6486 & 0.3370 & 0.0778 \\*
\rowcolor{white}
Stump & MGS & 0.7996 & 0.7806 & 0.6609 & 0.5194 & 0.3804 & 0.2914 \\*
\rowcolor{white}
Stump & ParetoGS & 0.8118 & 0.8158 & 0.7900 & 0.7532 & 0.6184 & 0.4473 \\
\SceneSeparator
\rowcolor{SceneTint}
Treehill & CLOD-3DGS & 0.6530 & 0.6311 & 0.5720 & 0.5187 & 0.2792 & 0.0628 \\*
\rowcolor{SceneTint}
Treehill & MGS & 0.6522 & 0.5995 & 0.4349 & 0.4035 & 0.3580 & 0.2586 \\*
\rowcolor{SceneTint}
Treehill & ParetoGS & 0.6562 & 0.6567 & 0.6238 & 0.5909 & 0.5009 & 0.4284 \\
\SceneSeparator
\rowcolor{white}
Room & CLOD-3DGS & 0.9301 & 0.9226 & 0.8938 & 0.8111 & 0.4076 & 0.1012 \\*
\rowcolor{white}
Room & MGS & 0.9328 & 0.9269 & 0.9026 & 0.8771 & 0.7796 & 0.5783 \\*
\rowcolor{white}
Room & ParetoGS & 0.9333 & 0.9292 & 0.9120 & 0.8965 & 0.8497 & 0.7641 \\
\SceneSeparator
\rowcolor{SceneTint}
Counter & CLOD-3DGS & 0.9165 & 0.9047 & 0.8526 & 0.7197 & 0.3606 & 0.0732 \\*
\rowcolor{SceneTint}
Counter & MGS & 0.9230 & 0.9151 & 0.8841 & 0.8465 & 0.7054 & 0.5473 \\*
\rowcolor{SceneTint}
Counter & ParetoGS & 0.9191 & 0.9152 & 0.8916 & 0.8673 & 0.7876 & 0.6857 \\
\SceneSeparator
\rowcolor{white}
Kitchen & CLOD-3DGS & 0.9333 & 0.9245 & 0.8869 & 0.7881 & 0.3920 & 0.0555 \\*
\rowcolor{white}
Kitchen & MGS & 0.9362 & 0.9300 & 0.8975 & 0.8621 & 0.6645 & 0.4471 \\*
\rowcolor{white}
Kitchen & ParetoGS & 0.9366 & 0.9290 & 0.8967 & 0.8649 & 0.7662 & 0.6251 \\
\SceneSeparator
\rowcolor{SceneTint}
Bonsai & CLOD-3DGS & 0.9473 & 0.9402 & 0.8896 & 0.7221 & 0.3717 & 0.0965 \\*
\rowcolor{SceneTint}
Bonsai & MGS & 0.9521 & 0.9471 & 0.9205 & 0.8798 & 0.7318 & 0.5622 \\*
\rowcolor{SceneTint}
Bonsai & ParetoGS & 0.9497 & 0.9468 & 0.9252 & 0.8999 & 0.8051 & 0.6921 \\
\SceneSeparator
\rowcolor{white}
Train & CLOD-3DGS & 0.8302 & 0.8007 & 0.6961 & 0.5333 & 0.2508 & 0.0381 \\*
\rowcolor{white}
Train & MGS & 0.8579 & 0.8390 & 0.7643 & 0.6902 & 0.5419 & 0.3913 \\*
\rowcolor{white}
Train & ParetoGS & 0.8442 & 0.8387 & 0.8010 & 0.7670 & 0.6695 & 0.5297 \\
\SceneSeparator
\rowcolor{SceneTint}
Truck & CLOD-3DGS & 0.9008 & 0.8913 & 0.8524 & 0.7876 & 0.4219 & 0.1086 \\*
\rowcolor{SceneTint}
Truck & MGS & 0.9022 & 0.8898 & 0.8389 & 0.7765 & 0.6173 & 0.4579 \\*
\rowcolor{SceneTint}
Truck & ParetoGS & 0.8991 & 0.8975 & 0.8789 & 0.8568 & 0.7716 & 0.6228 \\
\SceneSeparator
\rowcolor{white}
Playroom & CLOD-3DGS & 0.8964 & 0.9014 & 0.8923 & 0.8724 & 0.5805 & 0.0945 \\*
\rowcolor{white}
Playroom & MGS & 0.9110 & 0.9110 & 0.9043 & 0.8933 & 0.8070 & 0.6731 \\*
\rowcolor{white}
Playroom & ParetoGS & 0.9152 & 0.9095 & 0.9060 & 0.8996 & 0.8672 & 0.7975 \\
\SceneSeparator
\rowcolor{SceneTint}
DrJohnson & CLOD-3DGS & 0.9024 & 0.8982 & 0.8783 & 0.8477 & 0.6235 & 0.0828 \\*
\rowcolor{SceneTint}
DrJohnson & MGS & 0.9060 & 0.9044 & 0.8866 & 0.8662 & 0.7814 & 0.6652 \\*
\rowcolor{SceneTint}
DrJohnson & ParetoGS & 0.9122 & 0.9123 & 0.9065 & 0.8978 & 0.8608 & 0.7918 \\
\bottomrule
\end{longtable}
\endgroup

\begingroup
\setlength{\tabcolsep}{3pt}
\renewcommand{\arraystretch}{1.08}
\begin{longtable}{@{}llrrrrrr@{}}
\caption{Per-scene LPIPS at retention ratios from each method's own Full.}\label{tab:pareto-scenes-prefix-lpips} \\
\toprule
Scene & Method & 100\% & 30\% & 10\% & 5\% & 1\% & 0.1\% \\
\midrule
\endfirsthead
\toprule
Scene & Method & 100\% & 30\% & 10\% & 5\% & 1\% & 0.1\% \\
\midrule
\endhead
\rowcolor{SceneTint}
Bicycle & CLOD-3DGS & 0.1711 & 0.2078 & 0.3100 & 0.3859 & 0.5644 & 0.7261 \\*
\rowcolor{SceneTint}
Bicycle & MGS & 0.1670 & 0.2256 & 0.4185 & 0.5281 & 0.6185 & 0.6888 \\*
\rowcolor{SceneTint}
Bicycle & ParetoGS & 0.1613 & 0.1960 & 0.2872 & 0.3426 & 0.4634 & 0.5837 \\
\SceneSeparator
\rowcolor{white}
Flowers & CLOD-3DGS & 0.2877 & 0.3294 & 0.4124 & 0.4763 & 0.6536 & 0.8190 \\*
\rowcolor{white}
Flowers & MGS & 0.2659 & 0.3091 & 0.4390 & 0.5137 & 0.6453 & 0.7578 \\*
\rowcolor{white}
Flowers & ParetoGS & 0.2813 & 0.3322 & 0.4176 & 0.4712 & 0.5756 & 0.6911 \\
\SceneSeparator
\rowcolor{SceneTint}
Garden & CLOD-3DGS & 0.0988 & 0.1229 & 0.2223 & 0.3191 & 0.5634 & 0.7621 \\*
\rowcolor{SceneTint}
Garden & MGS & 0.0962 & 0.1308 & 0.2881 & 0.4218 & 0.5873 & 0.7057 \\*
\rowcolor{SceneTint}
Garden & ParetoGS & 0.0902 & 0.1135 & 0.2235 & 0.3008 & 0.4483 & 0.5849 \\
\SceneSeparator
\rowcolor{white}
Stump & CLOD-3DGS & 0.1838 & 0.2057 & 0.2869 & 0.3638 & 0.5539 & 0.6942 \\*
\rowcolor{white}
Stump & MGS & 0.1777 & 0.2068 & 0.3408 & 0.4709 & 0.6083 & 0.6700 \\*
\rowcolor{white}
Stump & ParetoGS & 0.1725 & 0.1865 & 0.2549 & 0.3137 & 0.4502 & 0.5970 \\
\SceneSeparator
\rowcolor{SceneTint}
Treehill & CLOD-3DGS & 0.2773 & 0.3219 & 0.4136 & 0.4786 & 0.6283 & 0.7154 \\*
\rowcolor{SceneTint}
Treehill & MGS & 0.2631 & 0.3352 & 0.5323 & 0.5775 & 0.6294 & 0.6916 \\*
\rowcolor{SceneTint}
Treehill & ParetoGS & 0.2816 & 0.3353 & 0.4156 & 0.4666 & 0.5573 & 0.6156 \\
\SceneSeparator
\rowcolor{white}
Room & CLOD-3DGS & 0.2003 & 0.2190 & 0.2710 & 0.3537 & 0.5718 & 0.6629 \\*
\rowcolor{white}
Room & MGS & 0.1905 & 0.2030 & 0.2474 & 0.2885 & 0.4063 & 0.5654 \\*
\rowcolor{white}
Room & ParetoGS & 0.1943 & 0.2074 & 0.2536 & 0.2863 & 0.3632 & 0.4648 \\
\SceneSeparator
\rowcolor{SceneTint}
Counter & CLOD-3DGS & 0.1879 & 0.2101 & 0.2765 & 0.3862 & 0.5728 & 0.6629 \\*
\rowcolor{SceneTint}
Counter & MGS & 0.1679 & 0.1807 & 0.2248 & 0.2735 & 0.4276 & 0.5731 \\*
\rowcolor{SceneTint}
Counter & ParetoGS & 0.1804 & 0.1933 & 0.2402 & 0.2789 & 0.3932 & 0.5146 \\
\SceneSeparator
\rowcolor{white}
Kitchen & CLOD-3DGS & 0.1234 & 0.1396 & 0.1959 & 0.2985 & 0.5886 & 0.7387 \\*
\rowcolor{white}
Kitchen & MGS & 0.1149 & 0.1258 & 0.1811 & 0.2287 & 0.4284 & 0.6099 \\*
\rowcolor{white}
Kitchen & ParetoGS & 0.1151 & 0.1307 & 0.1995 & 0.2523 & 0.3762 & 0.4890 \\
\SceneSeparator
\rowcolor{SceneTint}
Bonsai & CLOD-3DGS & 0.1956 & 0.2121 & 0.2833 & 0.4094 & 0.5685 & 0.6122 \\*
\rowcolor{SceneTint}
Bonsai & MGS & 0.1808 & 0.1905 & 0.2302 & 0.2830 & 0.4398 & 0.5628 \\*
\rowcolor{SceneTint}
Bonsai & ParetoGS & 0.1878 & 0.1982 & 0.2441 & 0.2862 & 0.4008 & 0.5154 \\
\SceneSeparator
\rowcolor{white}
Train & CLOD-3DGS & 0.1994 & 0.2410 & 0.3497 & 0.4719 & 0.6409 & 0.7028 \\*
\rowcolor{white}
Train & MGS & 0.1456 & 0.1709 & 0.2631 & 0.3466 & 0.4927 & 0.6255 \\*
\rowcolor{white}
Train & ParetoGS & 0.1711 & 0.1966 & 0.2634 & 0.3074 & 0.4139 & 0.5575 \\
\SceneSeparator
\rowcolor{SceneTint}
Truck & CLOD-3DGS & 0.1110 & 0.1352 & 0.1985 & 0.2726 & 0.5494 & 0.7092 \\*
\rowcolor{SceneTint}
Truck & MGS & 0.0956 & 0.1269 & 0.2221 & 0.3081 & 0.4735 & 0.6135 \\*
\rowcolor{SceneTint}
Truck & ParetoGS & 0.1179 & 0.1227 & 0.1712 & 0.2161 & 0.3352 & 0.4850 \\
\SceneSeparator
\rowcolor{white}
Playroom & CLOD-3DGS & 0.2719 & 0.2714 & 0.2971 & 0.3275 & 0.5848 & 0.8378 \\*
\rowcolor{white}
Playroom & MGS & 0.2333 & 0.2333 & 0.2548 & 0.2813 & 0.4072 & 0.5611 \\*
\rowcolor{white}
Playroom & ParetoGS & 0.2289 & 0.2412 & 0.2658 & 0.2881 & 0.3582 & 0.4566 \\
\SceneSeparator
\rowcolor{SceneTint}
DrJohnson & CLOD-3DGS & 0.2539 & 0.2673 & 0.3065 & 0.3489 & 0.5100 & 0.6082 \\*
\rowcolor{SceneTint}
DrJohnson & MGS & 0.2374 & 0.2413 & 0.2744 & 0.3099 & 0.4207 & 0.5331 \\*
\rowcolor{SceneTint}
DrJohnson & ParetoGS & 0.2280 & 0.2373 & 0.2617 & 0.2844 & 0.3543 & 0.4551 \\
\bottomrule
\end{longtable}
\endgroup

\begingroup
\setlength{\tabcolsep}{3pt}
\renewcommand{\arraystretch}{1.08}
\begin{longtable}{@{}llrrrrr@{}}
\caption{Per-scene PSNR (dB) at requested ratios from the shared Full.}\label{tab:pareto-scenes-shared_full-psnr} \\
\toprule
Scene & Method & 30\% & 10\% & 5\% & 1\% & 0.1\% \\
\midrule
\endfirsthead
\toprule
Scene & Method & 30\% & 10\% & 5\% & 1\% & 0.1\% \\
\midrule
\endhead
\rowcolor{SceneTint}
Bicycle & FlexGS-imp & 20.62 & 23.21 & 22.67 & 21.68 & 15.97 \\*
\rowcolor{SceneTint}
Bicycle & NanoGS-imp & 21.36 & 19.05 & 18.26 & 17.29 & 16.41 \\*
\rowcolor{SceneTint}
Bicycle & ParetoGS & 25.81 & 25.18 & 24.63 & 23.36 & 21.62 \\
\SceneSeparator
\rowcolor{white}
Flowers & FlexGS-imp & 16.99 & 19.45 & 18.87 & 18.00 & 14.41 \\*
\rowcolor{white}
Flowers & NanoGS-imp & 18.16 & 16.74 & 16.26 & 15.43 & 14.02 \\*
\rowcolor{white}
Flowers & ParetoGS & 21.84 & 21.23 & 20.66 & 19.25 & 17.72 \\
\SceneSeparator
\rowcolor{SceneTint}
Garden & FlexGS-imp & 21.24 & 24.06 & 22.89 & 21.40 & 12.82 \\*
\rowcolor{SceneTint}
Garden & NanoGS-imp & 22.51 & 20.12 & 19.15 & 18.06 & 16.06 \\*
\rowcolor{SceneTint}
Garden & ParetoGS & 27.93 & 26.65 & 25.70 & 23.64 & 21.52 \\
\SceneSeparator
\rowcolor{white}
Stump & FlexGS-imp & 17.64 & 23.55 & 22.38 & 21.07 & 12.17 \\*
\rowcolor{white}
Stump & NanoGS-imp & 23.15 & 20.59 & 19.89 & 19.03 & 17.71 \\*
\rowcolor{white}
Stump & ParetoGS & 27.31 & 26.90 & 26.34 & 24.47 & 21.98 \\
\SceneSeparator
\rowcolor{SceneTint}
Treehill & FlexGS-imp & 20.93 & 21.85 & 21.61 & 20.96 & 17.42 \\*
\rowcolor{SceneTint}
Treehill & NanoGS-imp & 19.50 & 18.32 & 18.03 & 17.73 & 16.50 \\*
\rowcolor{SceneTint}
Treehill & ParetoGS & 23.16 & 22.97 & 22.75 & 22.02 & 20.80 \\
\SceneSeparator
\rowcolor{white}
Room & FlexGS-imp & 25.31 & 29.99 & 28.47 & 24.95 & 17.47 \\*
\rowcolor{white}
Room & NanoGS-imp & 28.54 & 24.94 & 23.82 & 21.51 & 17.65 \\*
\rowcolor{white}
Room & ParetoGS & 32.55 & 31.84 & 31.13 & 28.61 & 23.36 \\
\SceneSeparator
\rowcolor{SceneTint}
Counter & FlexGS-imp & 25.03 & 27.33 & 25.94 & 22.72 & 15.76 \\*
\rowcolor{SceneTint}
Counter & NanoGS-imp & 25.98 & 23.32 & 22.05 & 20.03 & 16.73 \\*
\rowcolor{SceneTint}
Counter & ParetoGS & 29.66 & 28.86 & 28.03 & 25.53 & 21.37 \\
\SceneSeparator
\rowcolor{white}
Kitchen & FlexGS-imp & 21.59 & 27.31 & 25.36 & 22.49 & 14.47 \\*
\rowcolor{white}
Kitchen & NanoGS-imp & 26.37 & 23.20 & 21.83 & 19.49 & 16.65 \\*
\rowcolor{white}
Kitchen & ParetoGS & 32.15 & 30.28 & 28.68 & 25.17 & 21.17 \\
\SceneSeparator
\rowcolor{SceneTint}
Bonsai & FlexGS-imp & 26.81 & 29.35 & 27.37 & 23.25 & 18.00 \\*
\rowcolor{SceneTint}
Bonsai & NanoGS-imp & 27.72 & 23.98 & 22.74 & 20.70 & 17.14 \\*
\rowcolor{SceneTint}
Bonsai & ParetoGS & 32.89 & 31.28 & 29.81 & 26.30 & 22.27 \\
\SceneSeparator
\rowcolor{white}
Train & FlexGS-imp & 19.55 & 21.21 & 20.27 & 18.11 & 10.24 \\*
\rowcolor{white}
Train & NanoGS-imp & 18.81 & 16.48 & 15.33 & 13.89 & 12.59 \\*
\rowcolor{white}
Train & ParetoGS & 22.48 & 22.21 & 21.86 & 20.41 & 17.48 \\
\SceneSeparator
\rowcolor{SceneTint}
Truck & FlexGS-imp & 18.95 & 22.52 & 21.26 & 18.86 & 7.05 \\*
\rowcolor{SceneTint}
Truck & NanoGS-imp & 22.51 & 19.85 & 18.64 & 16.87 & 14.60 \\*
\rowcolor{SceneTint}
Truck & ParetoGS & 26.52 & 25.98 & 25.32 & 23.12 & 19.47 \\
\SceneSeparator
\rowcolor{white}
Playroom & FlexGS-imp & 25.45 & 29.28 & 27.80 & 24.32 & 8.96 \\*
\rowcolor{white}
Playroom & NanoGS-imp & 28.48 & 25.50 & 24.37 & 22.55 & 17.88 \\*
\rowcolor{white}
Playroom & ParetoGS & 30.32 & 30.26 & 30.00 & 27.69 & 22.41 \\
\SceneSeparator
\rowcolor{SceneTint}
DrJohnson & FlexGS-imp & 25.65 & 28.61 & 27.52 & 24.93 & 11.28 \\*
\rowcolor{SceneTint}
DrJohnson & NanoGS-imp & 27.12 & 24.62 & 24.01 & 22.38 & 18.84 \\*
\rowcolor{SceneTint}
DrJohnson & ParetoGS & 29.81 & 29.61 & 29.30 & 27.68 & 23.75 \\
\bottomrule
\end{longtable}
\endgroup

\begingroup
\setlength{\tabcolsep}{3pt}
\renewcommand{\arraystretch}{1.08}
\begin{longtable}{@{}llrrrrr@{}}
\caption{Per-scene SSIM at requested ratios from the shared Full.}\label{tab:pareto-scenes-shared_full-ssim} \\
\toprule
Scene & Method & 30\% & 10\% & 5\% & 1\% & 0.1\% \\
\midrule
\endfirsthead
\toprule
Scene & Method & 30\% & 10\% & 5\% & 1\% & 0.1\% \\
\midrule
\endhead
\rowcolor{SceneTint}
Bicycle & FlexGS-imp & 0.6193 & 0.6643 & 0.5950 & 0.4700 & 0.3109 \\*
\rowcolor{SceneTint}
Bicycle & NanoGS-imp & 0.5556 & 0.4102 & 0.3676 & 0.3247 & 0.2990 \\*
\rowcolor{SceneTint}
Bicycle & ParetoGS & 0.7988 & 0.7531 & 0.7072 & 0.5728 & 0.4388 \\
\SceneSeparator
\rowcolor{white}
Flowers & FlexGS-imp & 0.3761 & 0.4319 & 0.3909 & 0.3202 & 0.2314 \\*
\rowcolor{white}
Flowers & NanoGS-imp & 0.4279 & 0.3258 & 0.2924 & 0.2484 & 0.2221 \\*
\rowcolor{white}
Flowers & ParetoGS & 0.6272 & 0.5684 & 0.5165 & 0.3907 & 0.2859 \\
\SceneSeparator
\rowcolor{SceneTint}
Garden & FlexGS-imp & 0.6726 & 0.7067 & 0.6239 & 0.4766 & 0.2519 \\*
\rowcolor{SceneTint}
Garden & NanoGS-imp & 0.6025 & 0.4475 & 0.3925 & 0.3330 & 0.3014 \\*
\rowcolor{SceneTint}
Garden & ParetoGS & 0.8739 & 0.8207 & 0.7683 & 0.6039 & 0.4328 \\
\SceneSeparator
\rowcolor{white}
Stump & FlexGS-imp & 0.4418 & 0.6833 & 0.6000 & 0.4882 & 0.2000 \\*
\rowcolor{white}
Stump & NanoGS-imp & 0.6073 & 0.4548 & 0.4109 & 0.3654 & 0.3380 \\*
\rowcolor{white}
Stump & ParetoGS & 0.8158 & 0.7900 & 0.7532 & 0.6184 & 0.4473 \\
\SceneSeparator
\rowcolor{SceneTint}
Treehill & FlexGS-imp & 0.5795 & 0.5775 & 0.5301 & 0.4514 & 0.3680 \\*
\rowcolor{SceneTint}
Treehill & NanoGS-imp & 0.4555 & 0.3841 & 0.3719 & 0.3667 & 0.3511 \\*
\rowcolor{SceneTint}
Treehill & ParetoGS & 0.6567 & 0.6238 & 0.5909 & 0.5009 & 0.4284 \\
\SceneSeparator
\rowcolor{white}
Room & FlexGS-imp & 0.8308 & 0.8883 & 0.8592 & 0.7981 & 0.6636 \\*
\rowcolor{white}
Room & NanoGS-imp & 0.8521 & 0.7752 & 0.7473 & 0.6954 & 0.6246 \\*
\rowcolor{white}
Room & ParetoGS & 0.9292 & 0.9120 & 0.8965 & 0.8497 & 0.7641 \\
\SceneSeparator
\rowcolor{SceneTint}
Counter & FlexGS-imp & 0.8305 & 0.8561 & 0.8156 & 0.7335 & 0.5946 \\*
\rowcolor{SceneTint}
Counter & NanoGS-imp & 0.8173 & 0.7224 & 0.6824 & 0.6355 & 0.5755 \\*
\rowcolor{SceneTint}
Counter & ParetoGS & 0.9152 & 0.8916 & 0.8673 & 0.7876 & 0.6857 \\
\SceneSeparator
\rowcolor{white}
Kitchen & FlexGS-imp & 0.7763 & 0.8607 & 0.8133 & 0.6954 & 0.4645 \\*
\rowcolor{white}
Kitchen & NanoGS-imp & 0.8182 & 0.6871 & 0.6168 & 0.4960 & 0.4665 \\*
\rowcolor{white}
Kitchen & ParetoGS & 0.9290 & 0.8967 & 0.8649 & 0.7662 & 0.6251 \\
\SceneSeparator
\rowcolor{SceneTint}
Bonsai & FlexGS-imp & 0.8796 & 0.8926 & 0.8406 & 0.7313 & 0.6454 \\*
\rowcolor{SceneTint}
Bonsai & NanoGS-imp & 0.8647 & 0.7573 & 0.7124 & 0.6435 & 0.5692 \\*
\rowcolor{SceneTint}
Bonsai & ParetoGS & 0.9468 & 0.9252 & 0.8999 & 0.8051 & 0.6921 \\
\SceneSeparator
\rowcolor{white}
Train & FlexGS-imp & 0.7286 & 0.7465 & 0.6964 & 0.5945 & 0.3784 \\*
\rowcolor{white}
Train & NanoGS-imp & 0.6944 & 0.5696 & 0.5204 & 0.4638 & 0.4275 \\*
\rowcolor{white}
Train & ParetoGS & 0.8387 & 0.8010 & 0.7670 & 0.6695 & 0.5297 \\
\SceneSeparator
\rowcolor{SceneTint}
Truck & FlexGS-imp & 0.7413 & 0.7947 & 0.7492 & 0.6590 & 0.2311 \\*
\rowcolor{SceneTint}
Truck & NanoGS-imp & 0.7666 & 0.6462 & 0.5954 & 0.5250 & 0.4857 \\*
\rowcolor{SceneTint}
Truck & ParetoGS & 0.8975 & 0.8789 & 0.8568 & 0.7716 & 0.6228 \\
\SceneSeparator
\rowcolor{white}
Playroom & FlexGS-imp & 0.8499 & 0.8975 & 0.8778 & 0.8308 & 0.4138 \\*
\rowcolor{white}
Playroom & NanoGS-imp & 0.8786 & 0.8374 & 0.8167 & 0.7867 & 0.7323 \\*
\rowcolor{white}
Playroom & ParetoGS & 0.9095 & 0.9060 & 0.8996 & 0.8672 & 0.7975 \\
\SceneSeparator
\rowcolor{SceneTint}
DrJohnson & FlexGS-imp & 0.8403 & 0.8836 & 0.8618 & 0.8182 & 0.3451 \\*
\rowcolor{SceneTint}
DrJohnson & NanoGS-imp & 0.8679 & 0.8162 & 0.7991 & 0.7632 & 0.7135 \\*
\rowcolor{SceneTint}
DrJohnson & ParetoGS & 0.9123 & 0.9065 & 0.8978 & 0.8608 & 0.7918 \\
\bottomrule
\end{longtable}
\endgroup

\begingroup
\setlength{\tabcolsep}{3pt}
\renewcommand{\arraystretch}{1.08}
\begin{longtable}{@{}llrrrrr@{}}
\caption{Per-scene LPIPS at requested ratios from the shared Full.}\label{tab:pareto-scenes-shared_full-lpips} \\
\toprule
Scene & Method & 30\% & 10\% & 5\% & 1\% & 0.1\% \\
\midrule
\endfirsthead
\toprule
Scene & Method & 30\% & 10\% & 5\% & 1\% & 0.1\% \\
\midrule
\endhead
\rowcolor{SceneTint}
Bicycle & FlexGS-imp & 0.3459 & 0.3494 & 0.4160 & 0.5423 & 0.6872 \\*
\rowcolor{SceneTint}
Bicycle & NanoGS-imp & 0.3603 & 0.4801 & 0.5349 & 0.6497 & 0.7321 \\*
\rowcolor{SceneTint}
Bicycle & ParetoGS & 0.1960 & 0.2872 & 0.3426 & 0.4634 & 0.5837 \\
\SceneSeparator
\rowcolor{white}
Flowers & FlexGS-imp & 0.5558 & 0.5246 & 0.5619 & 0.6399 & 0.7682 \\*
\rowcolor{white}
Flowers & NanoGS-imp & 0.4562 & 0.5526 & 0.5984 & 0.7128 & 0.7745 \\*
\rowcolor{white}
Flowers & ParetoGS & 0.3322 & 0.4176 & 0.4712 & 0.5756 & 0.6911 \\
\SceneSeparator
\rowcolor{SceneTint}
Garden & FlexGS-imp & 0.2943 & 0.3038 & 0.3894 & 0.5302 & 0.7072 \\*
\rowcolor{SceneTint}
Garden & NanoGS-imp & 0.3117 & 0.4373 & 0.4970 & 0.6157 & 0.7597 \\*
\rowcolor{SceneTint}
Garden & ParetoGS & 0.1135 & 0.2235 & 0.3008 & 0.4483 & 0.5849 \\
\SceneSeparator
\rowcolor{white}
Stump & FlexGS-imp & 0.5328 & 0.3311 & 0.4151 & 0.5242 & 0.6659 \\*
\rowcolor{white}
Stump & NanoGS-imp & 0.3406 & 0.4700 & 0.5249 & 0.6337 & 0.7089 \\*
\rowcolor{white}
Stump & ParetoGS & 0.1865 & 0.2549 & 0.3137 & 0.4502 & 0.5970 \\
\SceneSeparator
\rowcolor{SceneTint}
Treehill & FlexGS-imp & 0.4079 & 0.4555 & 0.5175 & 0.5937 & 0.6653 \\*
\rowcolor{SceneTint}
Treehill & NanoGS-imp & 0.4885 & 0.5723 & 0.6106 & 0.6815 & 0.6710 \\*
\rowcolor{SceneTint}
Treehill & ParetoGS & 0.3353 & 0.4156 & 0.4666 & 0.5573 & 0.6156 \\
\SceneSeparator
\rowcolor{white}
Room & FlexGS-imp & 0.3311 & 0.2920 & 0.3429 & 0.4254 & 0.5183 \\*
\rowcolor{white}
Room & NanoGS-imp & 0.2986 & 0.3825 & 0.4223 & 0.5020 & 0.5375 \\*
\rowcolor{white}
Room & ParetoGS & 0.2074 & 0.2536 & 0.2863 & 0.3632 & 0.4648 \\
\SceneSeparator
\rowcolor{SceneTint}
Counter & FlexGS-imp & 0.2919 & 0.2871 & 0.3450 & 0.4448 & 0.5561 \\*
\rowcolor{SceneTint}
Counter & NanoGS-imp & 0.2896 & 0.3913 & 0.4388 & 0.5256 & 0.5730 \\*
\rowcolor{SceneTint}
Counter & ParetoGS & 0.1933 & 0.2402 & 0.2789 & 0.3932 & 0.5146 \\
\SceneSeparator
\rowcolor{white}
Kitchen & FlexGS-imp & 0.2961 & 0.2499 & 0.3097 & 0.4332 & 0.6270 \\*
\rowcolor{white}
Kitchen & NanoGS-imp & 0.2431 & 0.3752 & 0.4421 & 0.5730 & 0.6484 \\*
\rowcolor{white}
Kitchen & ParetoGS & 0.1307 & 0.1995 & 0.2523 & 0.3762 & 0.4890 \\
\SceneSeparator
\rowcolor{SceneTint}
Bonsai & FlexGS-imp & 0.2757 & 0.2876 & 0.3487 & 0.4632 & 0.5358 \\*
\rowcolor{SceneTint}
Bonsai & NanoGS-imp & 0.2864 & 0.3970 & 0.4454 & 0.5261 & 0.5716 \\*
\rowcolor{SceneTint}
Bonsai & ParetoGS & 0.1982 & 0.2441 & 0.2862 & 0.4008 & 0.5154 \\
\SceneSeparator
\rowcolor{white}
Train & FlexGS-imp & 0.2961 & 0.3084 & 0.3657 & 0.4804 & 0.6470 \\*
\rowcolor{white}
Train & NanoGS-imp & 0.3228 & 0.4530 & 0.5119 & 0.5987 & 0.6335 \\*
\rowcolor{white}
Train & ParetoGS & 0.1966 & 0.2634 & 0.3074 & 0.4139 & 0.5575 \\
\SceneSeparator
\rowcolor{SceneTint}
Truck & FlexGS-imp & 0.2898 & 0.2660 & 0.3291 & 0.4435 & 0.6905 \\*
\rowcolor{SceneTint}
Truck & NanoGS-imp & 0.2659 & 0.3960 & 0.4596 & 0.5691 & 0.6572 \\*
\rowcolor{SceneTint}
Truck & ParetoGS & 0.1227 & 0.1712 & 0.2161 & 0.3352 & 0.4850 \\
\SceneSeparator
\rowcolor{white}
Playroom & FlexGS-imp & 0.3436 & 0.2968 & 0.3355 & 0.4114 & 0.6926 \\*
\rowcolor{white}
Playroom & NanoGS-imp & 0.2975 & 0.3664 & 0.3989 & 0.4557 & 0.4982 \\*
\rowcolor{white}
Playroom & ParetoGS & 0.2412 & 0.2658 & 0.2881 & 0.3582 & 0.4566 \\
\SceneSeparator
\rowcolor{SceneTint}
DrJohnson & FlexGS-imp & 0.3518 & 0.3017 & 0.3390 & 0.4095 & 0.5738 \\*
\rowcolor{SceneTint}
DrJohnson & NanoGS-imp & 0.3145 & 0.4002 & 0.4353 & 0.4872 & 0.5196 \\*
\rowcolor{SceneTint}
DrJohnson & ParetoGS & 0.2373 & 0.2617 & 0.2844 & 0.3543 & 0.4551 \\
\bottomrule
\end{longtable}
\endgroup

\begingroup
\setlength{\tabcolsep}{3pt}
\renewcommand{\arraystretch}{1.08}
\begin{longtable}{@{}llrrrrrr@{}}
\caption{Per-scene actual Gaussian counts (K) at retention ratios from each method's own Full.}\label{tab:pareto-scenes-prefix-count} \\
\toprule
Scene & Method & 100\% & 30\% & 10\% & 5\% & 1\% & 0.1\% \\
\midrule
\endfirsthead
\toprule
Scene & Method & 100\% & 30\% & 10\% & 5\% & 1\% & 0.1\% \\
\midrule
\endhead
\rowcolor{SceneTint}
Bicycle & CLOD-3DGS & 6132 & 1840 & 613 & 307 & 61 & 6 \\*
\rowcolor{SceneTint}
Bicycle & MGS & 5000 & 1500 & 500 & 250 & 50 & 5 \\*
\rowcolor{SceneTint}
Bicycle & ParetoGS & 3000 & 900 & 300 & 150 & 30 & 3 \\
\SceneSeparator
\rowcolor{white}
Flowers & CLOD-3DGS & 3636 & 1091 & 364 & 182 & 36 & 4 \\*
\rowcolor{white}
Flowers & MGS & 5000 & 1500 & 500 & 250 & 50 & 5 \\*
\rowcolor{white}
Flowers & ParetoGS & 1500 & 450 & 150 & 75 & 15 & 2 \\
\SceneSeparator
\rowcolor{SceneTint}
Garden & CLOD-3DGS & 5835 & 1750 & 583 & 292 & 58 & 6 \\*
\rowcolor{SceneTint}
Garden & MGS & 5000 & 1500 & 500 & 250 & 50 & 5 \\*
\rowcolor{SceneTint}
Garden & ParetoGS & 3000 & 900 & 300 & 150 & 30 & 3 \\
\SceneSeparator
\rowcolor{white}
Stump & CLOD-3DGS & 4962 & 1489 & 496 & 248 & 50 & 5 \\*
\rowcolor{white}
Stump & MGS & 5000 & 1500 & 500 & 250 & 50 & 5 \\*
\rowcolor{white}
Stump & ParetoGS & 3000 & 900 & 300 & 150 & 30 & 3 \\
\SceneSeparator
\rowcolor{SceneTint}
Treehill & CLOD-3DGS & 3784 & 1135 & 378 & 189 & 38 & 4 \\*
\rowcolor{SceneTint}
Treehill & MGS & 5000 & 1500 & 500 & 250 & 50 & 5 \\*
\rowcolor{SceneTint}
Treehill & ParetoGS & 1500 & 450 & 150 & 75 & 15 & 2 \\
\SceneSeparator
\rowcolor{white}
Room & CLOD-3DGS & 1593 & 478 & 159 & 80 & 16 & 2 \\*
\rowcolor{white}
Room & MGS & 5000 & 1500 & 500 & 250 & 50 & 5 \\*
\rowcolor{white}
Room & ParetoGS & 1000 & 300 & 100 & 50 & 10 & 1 \\
\SceneSeparator
\rowcolor{SceneTint}
Counter & CLOD-3DGS & 1223 & 367 & 122 & 61 & 12 & 1 \\*
\rowcolor{SceneTint}
Counter & MGS & 5000 & 1500 & 500 & 250 & 50 & 5 \\*
\rowcolor{SceneTint}
Counter & ParetoGS & 1000 & 300 & 100 & 50 & 10 & 1 \\
\SceneSeparator
\rowcolor{white}
Kitchen & CLOD-3DGS & 1852 & 556 & 185 & 93 & 19 & 2 \\*
\rowcolor{white}
Kitchen & MGS & 5000 & 1500 & 500 & 250 & 50 & 5 \\*
\rowcolor{white}
Kitchen & ParetoGS & 1000 & 300 & 100 & 50 & 10 & 1 \\
\SceneSeparator
\rowcolor{SceneTint}
Bonsai & CLOD-3DGS & 1245 & 373 & 124 & 62 & 12 & 1 \\*
\rowcolor{SceneTint}
Bonsai & MGS & 5000 & 1500 & 500 & 250 & 50 & 5 \\*
\rowcolor{SceneTint}
Bonsai & ParetoGS & 1000 & 300 & 100 & 50 & 10 & 1 \\
\SceneSeparator
\rowcolor{white}
Train & CLOD-3DGS & 1027 & 308 & 103 & 51 & 10 & 1 \\*
\rowcolor{white}
Train & MGS & 5000 & 1500 & 500 & 250 & 50 & 5 \\*
\rowcolor{white}
Train & ParetoGS & 1000 & 300 & 100 & 50 & 10 & 1 \\
\SceneSeparator
\rowcolor{SceneTint}
Truck & CLOD-3DGS & 2541 & 762 & 254 & 127 & 25 & 3 \\*
\rowcolor{SceneTint}
Truck & MGS & 5000 & 1500 & 500 & 250 & 50 & 5 \\*
\rowcolor{SceneTint}
Truck & ParetoGS & 1500 & 450 & 150 & 75 & 15 & 2 \\
\SceneSeparator
\rowcolor{white}
Playroom & CLOD-3DGS & 2546 & 764 & 255 & 127 & 25 & 3 \\*
\rowcolor{white}
Playroom & MGS & 5000 & 1500 & 500 & 250 & 50 & 5 \\*
\rowcolor{white}
Playroom & ParetoGS & 1000 & 300 & 100 & 50 & 10 & 1 \\
\SceneSeparator
\rowcolor{SceneTint}
DrJohnson & CLOD-3DGS & 3000 & 900 & 300 & 150 & 30 & 3 \\*
\rowcolor{SceneTint}
DrJohnson & MGS & 4000 & 1200 & 400 & 200 & 40 & 4 \\*
\rowcolor{SceneTint}
DrJohnson & ParetoGS & 1500 & 450 & 150 & 75 & 15 & 2 \\
\bottomrule
\end{longtable}
\endgroup

\begingroup
\setlength{\tabcolsep}{3pt}
\renewcommand{\arraystretch}{1.08}
\begin{longtable}{@{}llrrrrr@{}}
\caption{Per-scene requested and FlexGS-imp actual output counts (K) from the shared Full.}\label{tab:pareto-scenes-shared-count} \\
\toprule
Scene & Count & 30\% & 10\% & 5\% & 1\% & 0.1\% \\
\midrule
\endfirsthead
\toprule
Scene & Count & 30\% & 10\% & 5\% & 1\% & 0.1\% \\
\midrule
\endhead
\rowcolor{SceneTint}
Bicycle & Requested & 900 & 300 & 150 & 30 & 3 \\*
\rowcolor{SceneTint}
Bicycle & FlexGS actual & 900 & 300 & 150 & 30 & 3 \\
\SceneSeparator
\rowcolor{white}
Flowers & Requested & 450 & 150 & 75 & 15 & 2 \\*
\rowcolor{white}
Flowers & FlexGS actual & 450 & 150 & 75 & 15 & 2 \\
\SceneSeparator
\rowcolor{SceneTint}
Garden & Requested & 900 & 300 & 150 & 30 & 3 \\*
\rowcolor{SceneTint}
Garden & FlexGS actual & 900 & 300 & 150 & 30 & 3 \\
\SceneSeparator
\rowcolor{white}
Stump & Requested & 900 & 300 & 150 & 30 & 3 \\*
\rowcolor{white}
Stump & FlexGS actual & 900 & 300 & 150 & 30 & 3 \\
\SceneSeparator
\rowcolor{SceneTint}
Treehill & Requested & 450 & 150 & 75 & 15 & 2 \\*
\rowcolor{SceneTint}
Treehill & FlexGS actual & 450 & 150 & 75 & 15 & 3 \\
\SceneSeparator
\rowcolor{white}
Room & Requested & 300 & 100 & 50 & 10 & 1 \\*
\rowcolor{white}
Room & FlexGS actual & 300 & 100 & 50 & 10 & 3 \\
\SceneSeparator
\rowcolor{SceneTint}
Counter & Requested & 300 & 100 & 50 & 10 & 1 \\*
\rowcolor{SceneTint}
Counter & FlexGS actual & 300 & 100 & 50 & 10 & 3 \\
\SceneSeparator
\rowcolor{white}
Kitchen & Requested & 300 & 100 & 50 & 10 & 1 \\*
\rowcolor{white}
Kitchen & FlexGS actual & 300 & 100 & 50 & 10 & 1 \\
\SceneSeparator
\rowcolor{SceneTint}
Bonsai & Requested & 300 & 100 & 50 & 10 & 1 \\*
\rowcolor{SceneTint}
Bonsai & FlexGS actual & 300 & 100 & 50 & 10 & 5 \\
\SceneSeparator
\rowcolor{white}
Train & Requested & 300 & 100 & 50 & 10 & 1 \\*
\rowcolor{white}
Train & FlexGS actual & 300 & 100 & 50 & 10 & 1 \\
\SceneSeparator
\rowcolor{SceneTint}
Truck & Requested & 450 & 150 & 75 & 15 & 2 \\*
\rowcolor{SceneTint}
Truck & FlexGS actual & 450 & 150 & 75 & 15 & 2 \\
\SceneSeparator
\rowcolor{white}
Playroom & Requested & 300 & 100 & 50 & 10 & 1 \\*
\rowcolor{white}
Playroom & FlexGS actual & 300 & 100 & 50 & 10 & 1 \\
\SceneSeparator
\rowcolor{SceneTint}
DrJohnson & Requested & 450 & 150 & 75 & 15 & 2 \\*
\rowcolor{SceneTint}
DrJohnson & FlexGS actual & 450 & 150 & 75 & 15 & 2 \\
\bottomrule
\end{longtable}
\endgroup

\begingroup
\setlength{\tabcolsep}{3pt}
\renewcommand{\arraystretch}{1.08}
\begin{longtable}{@{}lrrrrr@{}}
\caption{Per-scene NanoGS-imp offline conversion time (seconds) from the shared Full to each target PLY.}\label{tab:pareto-scenes-nano-conv} \\
\toprule
Scene & 30\% & 10\% & 5\% & 1\% & 0.1\% \\
\midrule
\endfirsthead
\toprule
Scene & 30\% & 10\% & 5\% & 1\% & 0.1\% \\
\midrule
\endhead
\rowcolor{SceneTint}
Bicycle & 99 & 108 & 110 & 111 & 110 \\
\SceneSeparator
\rowcolor{white}
Flowers & 49 & 57 & 57 & 58 & 59 \\
\SceneSeparator
\rowcolor{SceneTint}
Garden & 96 & 106 & 111 & 110 & 110 \\
\SceneSeparator
\rowcolor{white}
Stump & 87 & 101 & 103 & 104 & 104 \\
\SceneSeparator
\rowcolor{SceneTint}
Treehill & 49 & 56 & 57 & 58 & 59 \\
\SceneSeparator
\rowcolor{white}
Room & 28 & 32 & 32 & 33 & 34 \\
\SceneSeparator
\rowcolor{SceneTint}
Counter & 27 & 30 & 31 & 31 & 31 \\
\SceneSeparator
\rowcolor{white}
Kitchen & 28 & 31 & 32 & 32 & 32 \\
\SceneSeparator
\rowcolor{SceneTint}
Bonsai & 26 & 27 & 29 & 30 & 30 \\
\SceneSeparator
\rowcolor{white}
Train & 28 & 31 & 32 & 32 & 32 \\
\SceneSeparator
\rowcolor{SceneTint}
Truck & 37 & 39 & 41 & 42 & 42 \\
\SceneSeparator
\rowcolor{white}
Playroom & 29 & 33 & 33 & 34 & 34 \\
\SceneSeparator
\rowcolor{SceneTint}
DrJohnson & 46 & 52 & 53 & 54 & 54 \\
\bottomrule
\end{longtable}
\endgroup

\begingroup
\setlength{\tabcolsep}{3pt}
\renewcommand{\arraystretch}{1.08}
\begin{longtable}{@{}lrrrrrrrr@{}}
\caption{Per-scene offline time. CLOD, MGS, FlexGS, ParetoGS training, and total are in minutes; encoding, calibration, and packaging are in seconds. ParetoGS total includes its training and all three post-processing stages.}\label{tab:pareto-scenes-time} \\
\toprule
Scene & CLOD & MGS & FlexGS & Training & Encoding & Calibration & Packaging & Total \\
\midrule
\endfirsthead
\toprule
Scene & CLOD & MGS & FlexGS & Training & Encoding & Calibration & Packaging & Total \\
\midrule
\endhead
\rowcolor{SceneTint}
Bicycle & 57.24 & 68.74 & 120.18 & 12.28 & 1.98 & 31.91 & 3.98 & 12.92 \\
\SceneSeparator
\rowcolor{white}
Flowers & 38.10 & 71.78 & 74.74 & 9.29 & 1.60 & 26.29 & 2.43 & 9.80 \\
\SceneSeparator
\rowcolor{SceneTint}
Garden & 56.35 & 72.58 & 123.04 & 13.53 & 1.83 & 33.10 & 3.86 & 14.18 \\
\SceneSeparator
\rowcolor{white}
Stump & 46.94 & 68.18 & 119.88 & 12.12 & 1.74 & 29.84 & 3.75 & 12.71 \\
\SceneSeparator
\rowcolor{SceneTint}
Treehill & 38.46 & 68.18 & 72.29 & 8.79 & 2.07 & 24.74 & 2.41 & 9.27 \\
\SceneSeparator
\rowcolor{white}
Room & 29.36 & 82.03 & 69.89 & 9.62 & 1.43 & 33.23 & 1.91 & 10.23 \\
\SceneSeparator
\rowcolor{SceneTint}
Counter & 27.92 & 89.00 & 70.15 & 10.91 & 1.28 & 33.93 & 1.90 & 11.53 \\
\SceneSeparator
\rowcolor{white}
Kitchen & 32.91 & 91.10 & 73.24 & 10.51 & 1.35 & 34.16 & 1.88 & 11.14 \\
\SceneSeparator
\rowcolor{SceneTint}
Bonsai & 24.78 & 85.42 & 72.10 & 9.85 & 1.29 & 33.78 & 1.89 & 10.46 \\
\SceneSeparator
\rowcolor{white}
Train & 14.16 & 76.40 & 56.66 & 6.84 & 1.29 & 22.17 & 1.91 & 7.26 \\
\SceneSeparator
\rowcolor{SceneTint}
Truck & 27.51 & 75.08 & 74.65 & 8.01 & 1.47 & 23.76 & 2.39 & 8.47 \\
\SceneSeparator
\rowcolor{white}
Playroom & 27.67 & 64.02 & 57.02 & 7.08 & 1.24 & 22.62 & 1.90 & 7.51 \\
\SceneSeparator
\rowcolor{SceneTint}
DrJohnson & 36.90 & 54.52 & 73.28 & 8.31 & 1.43 & 26.94 & 2.48 & 8.83 \\
\SceneSeparator
\rowcolor{white}
13-scene mean & 35.25 & 74.39 & 81.32 & 9.78 & 1.54 & 28.96 & 2.51 & 10.33 \\
\bottomrule
\end{longtable}
\endgroup

%% file: tables/decoding_scene.tex

\begingroup
\setlength{\tabcolsep}{3pt}
\renewcommand{\arraystretch}{1.08}
\begin{longtable}{@{}llrrr@{}}
\caption{Per-scene results for Table~\ref{tab:pareto-decoding}. Each scene averages ten interval midpoints; the Full-parameter row uses the same member IDs.}\label{tab:pareto-decoding-scenes} \\
\toprule
Scene & Variant & PSNR$\uparrow$ & SSIM$\uparrow$ & LPIPS$\downarrow$ \\
\midrule
\endfirsthead
\toprule
Scene & Variant & PSNR$\uparrow$ & SSIM$\uparrow$ & LPIPS$\downarrow$ \\
\midrule
\endhead
\rowcolor{SceneTint}
Bicycle & Full method & 24.16 & 0.6544 & 0.3687 \\*
\rowcolor{SceneTint}
Bicycle & Fixed shared state & 24.04 & 0.6523 & 0.3701 \\*
\rowcolor{SceneTint}
Bicycle & No opacity gate & 24.03 & 0.6522 & 0.3686 \\*
\rowcolor{SceneTint}
Bicycle & Linear progress & 24.08 & 0.6494 & 0.3732 \\*
\rowcolor{SceneTint}
Bicycle & Full parameters & 15.57 & 0.3871 & 0.4946 \\
\SceneSeparator
\rowcolor{white}
Flowers & Full method & 20.12 & 0.4789 & 0.4894 \\*
\rowcolor{white}
Flowers & Fixed shared state & 20.07 & 0.4766 & 0.4915 \\*
\rowcolor{white}
Flowers & No opacity gate & 20.03 & 0.4779 & 0.4889 \\*
\rowcolor{white}
Flowers & Linear progress & 20.05 & 0.4732 & 0.4943 \\*
\rowcolor{white}
Flowers & Full parameters & 13.84 & 0.3186 & 0.5860 \\
\SceneSeparator
\rowcolor{SceneTint}
Garden & Full method & 25.13 & 0.7012 & 0.3286 \\*
\rowcolor{SceneTint}
Garden & Fixed shared state & 24.94 & 0.6988 & 0.3300 \\*
\rowcolor{SceneTint}
Garden & No opacity gate & 24.81 & 0.6971 & 0.3294 \\*
\rowcolor{SceneTint}
Garden & Linear progress & 24.96 & 0.6948 & 0.3333 \\*
\rowcolor{SceneTint}
Garden & Full parameters & 14.81 & 0.4177 & 0.4917 \\
\SceneSeparator
\rowcolor{white}
Stump & Full method & 25.43 & 0.6865 & 0.3560 \\*
\rowcolor{white}
Stump & Fixed shared state & 25.33 & 0.6841 & 0.3582 \\*
\rowcolor{white}
Stump & No opacity gate & 25.26 & 0.6837 & 0.3562 \\*
\rowcolor{white}
Stump & Linear progress & 25.30 & 0.6808 & 0.3606 \\*
\rowcolor{white}
Stump & Full parameters & 17.62 & 0.4127 & 0.4713 \\
\SceneSeparator
\rowcolor{SceneTint}
Treehill & Full method & 22.32 & 0.5597 & 0.4716 \\*
\rowcolor{SceneTint}
Treehill & Fixed shared state & 22.31 & 0.5585 & 0.4731 \\*
\rowcolor{SceneTint}
Treehill & No opacity gate & 22.25 & 0.5578 & 0.4718 \\*
\rowcolor{SceneTint}
Treehill & Linear progress & 22.22 & 0.5558 & 0.4749 \\*
\rowcolor{SceneTint}
Treehill & Full parameters & 15.14 & 0.3871 & 0.5585 \\
\SceneSeparator
\rowcolor{white}
Room & Full method & 29.59 & 0.8723 & 0.3118 \\*
\rowcolor{white}
Room & Fixed shared state & 29.40 & 0.8710 & 0.3136 \\*
\rowcolor{white}
Room & No opacity gate & 29.36 & 0.8707 & 0.3116 \\*
\rowcolor{white}
Room & Linear progress & 29.43 & 0.8700 & 0.3138 \\*
\rowcolor{white}
Room & Full parameters & 20.30 & 0.7013 & 0.4290 \\
\SceneSeparator
\rowcolor{SceneTint}
Counter & Full method & 26.75 & 0.8311 & 0.3217 \\*
\rowcolor{SceneTint}
Counter & Fixed shared state & 26.60 & 0.8294 & 0.3241 \\*
\rowcolor{SceneTint}
Counter & No opacity gate & 26.61 & 0.8297 & 0.3211 \\*
\rowcolor{SceneTint}
Counter & Linear progress & 26.61 & 0.8280 & 0.3243 \\*
\rowcolor{SceneTint}
Counter & Full parameters & 18.58 & 0.6393 & 0.4390 \\
\SceneSeparator
\rowcolor{white}
Kitchen & Full method & 27.57 & 0.8202 & 0.2850 \\*
\rowcolor{white}
Kitchen & Fixed shared state & 27.33 & 0.8177 & 0.2873 \\*
\rowcolor{white}
Kitchen & No opacity gate & 26.75 & 0.8158 & 0.2856 \\*
\rowcolor{white}
Kitchen & Linear progress & 27.21 & 0.8164 & 0.2865 \\*
\rowcolor{white}
Kitchen & Full parameters & 15.57 & 0.5316 & 0.4713 \\
\SceneSeparator
\rowcolor{SceneTint}
Bonsai & Full method & 28.59 & 0.8550 & 0.3261 \\*
\rowcolor{SceneTint}
Bonsai & Fixed shared state & 28.41 & 0.8534 & 0.3277 \\*
\rowcolor{SceneTint}
Bonsai & No opacity gate & 28.21 & 0.8531 & 0.3256 \\*
\rowcolor{SceneTint}
Bonsai & Linear progress & 28.36 & 0.8515 & 0.3290 \\*
\rowcolor{SceneTint}
Bonsai & Full parameters & 19.32 & 0.6524 & 0.4391 \\
\SceneSeparator
\rowcolor{white}
Train & Full method & 20.93 & 0.7245 & 0.3407 \\*
\rowcolor{white}
Train & Fixed shared state & 20.87 & 0.7224 & 0.3429 \\*
\rowcolor{white}
Train & No opacity gate & 20.82 & 0.7227 & 0.3402 \\*
\rowcolor{white}
Train & Linear progress & 20.84 & 0.7207 & 0.3437 \\*
\rowcolor{white}
Train & Full parameters & 14.66 & 0.5443 & 0.4630 \\
\SceneSeparator
\rowcolor{SceneTint}
Truck & Full method & 24.12 & 0.8090 & 0.2624 \\*
\rowcolor{SceneTint}
Truck & Fixed shared state & 23.99 & 0.8068 & 0.2652 \\*
\rowcolor{SceneTint}
Truck & No opacity gate & 23.97 & 0.8075 & 0.2619 \\*
\rowcolor{SceneTint}
Truck & Linear progress & 24.00 & 0.8050 & 0.2662 \\*
\rowcolor{SceneTint}
Truck & Full parameters & 16.54 & 0.6110 & 0.4096 \\
\SceneSeparator
\rowcolor{white}
Playroom & Full method & 28.29 & 0.8781 & 0.3191 \\*
\rowcolor{white}
Playroom & Fixed shared state & 28.17 & 0.8773 & 0.3210 \\*
\rowcolor{white}
Playroom & No opacity gate & 28.22 & 0.8777 & 0.3186 \\*
\rowcolor{white}
Playroom & Linear progress & 28.15 & 0.8764 & 0.3214 \\*
\rowcolor{white}
Playroom & Full parameters & 19.89 & 0.7582 & 0.4398 \\
\SceneSeparator
\rowcolor{SceneTint}
DrJohnson & Full method & 28.16 & 0.8762 & 0.3141 \\*
\rowcolor{SceneTint}
DrJohnson & Fixed shared state & 28.02 & 0.8751 & 0.3158 \\*
\rowcolor{SceneTint}
DrJohnson & No opacity gate & 28.06 & 0.8754 & 0.3137 \\*
\rowcolor{SceneTint}
DrJohnson & Linear progress & 28.07 & 0.8746 & 0.3166 \\*
\rowcolor{SceneTint}
DrJohnson & Full parameters & 20.12 & 0.7351 & 0.4399 \\
\bottomrule
\end{longtable}
\endgroup

%% file: tables/storage_scene.tex

\begingroup
\setlength{\tabcolsep}{3pt}
\renewcommand{\arraystretch}{1.08}
\begin{longtable}{@{}llrrrr@{}}
\caption{Per-scene results for Table~\ref{tab:pareto-storage}. Quality averages ten interval midpoints per scene; MB is additional state storage.}\label{tab:pareto-storage-scenes} \\
\toprule
Scene & Encoding & PSNR$\uparrow$ & SSIM$\uparrow$ & LPIPS$\downarrow$ & MB$\downarrow$ \\
\midrule
\endfirsthead
\toprule
Scene & Encoding & PSNR$\uparrow$ & SSIM$\uparrow$ & LPIPS$\downarrow$ & MB$\downarrow$ \\
\midrule
\endhead
\rowcolor{SceneTint}
Bicycle & FP32 & 24.18 & 0.6550 & 0.3678 & 724.3 \\*
\rowcolor{SceneTint}
Bicycle & FP16 & 24.18 & 0.6550 & 0.3678 & 362.2 \\*
\rowcolor{SceneTint}
Bicycle & Atlas & 24.16 & 0.6544 & 0.3687 & 158.7 \\
\SceneSeparator
\rowcolor{white}
Flowers & FP32 & 20.13 & 0.4792 & 0.4890 & 362.2 \\*
\rowcolor{white}
Flowers & FP16 & 20.13 & 0.4792 & 0.4890 & 181.1 \\*
\rowcolor{white}
Flowers & Atlas & 20.12 & 0.4789 & 0.4894 & 79.6 \\
\SceneSeparator
\rowcolor{SceneTint}
Garden & FP32 & 25.15 & 0.7016 & 0.3278 & 724.3 \\*
\rowcolor{SceneTint}
Garden & FP16 & 25.15 & 0.7016 & 0.3278 & 362.2 \\*
\rowcolor{SceneTint}
Garden & Atlas & 25.13 & 0.7012 & 0.3286 & 158.7 \\
\SceneSeparator
\rowcolor{white}
Stump & FP32 & 25.44 & 0.6869 & 0.3551 & 724.3 \\*
\rowcolor{white}
Stump & FP16 & 25.44 & 0.6869 & 0.3551 & 362.2 \\*
\rowcolor{white}
Stump & Atlas & 25.43 & 0.6865 & 0.3560 & 158.7 \\
\SceneSeparator
\rowcolor{SceneTint}
Treehill & FP32 & 22.32 & 0.5600 & 0.4711 & 362.2 \\*
\rowcolor{SceneTint}
Treehill & FP16 & 22.32 & 0.5600 & 0.4711 & 181.1 \\*
\rowcolor{SceneTint}
Treehill & Atlas & 22.32 & 0.5597 & 0.4716 & 79.6 \\
\SceneSeparator
\rowcolor{white}
Room & FP32 & 29.60 & 0.8726 & 0.3116 & 241.4 \\*
\rowcolor{white}
Room & FP16 & 29.60 & 0.8726 & 0.3116 & 120.7 \\*
\rowcolor{white}
Room & Atlas & 29.59 & 0.8723 & 0.3118 & 53.2 \\
\SceneSeparator
\rowcolor{SceneTint}
Counter & FP32 & 26.76 & 0.8315 & 0.3214 & 241.4 \\*
\rowcolor{SceneTint}
Counter & FP16 & 26.76 & 0.8315 & 0.3214 & 120.7 \\*
\rowcolor{SceneTint}
Counter & Atlas & 26.75 & 0.8311 & 0.3217 & 53.2 \\
\SceneSeparator
\rowcolor{white}
Kitchen & FP32 & 27.60 & 0.8220 & 0.2839 & 241.4 \\*
\rowcolor{white}
Kitchen & FP16 & 27.60 & 0.8220 & 0.2839 & 120.7 \\*
\rowcolor{white}
Kitchen & Atlas & 27.57 & 0.8202 & 0.2850 & 53.2 \\
\SceneSeparator
\rowcolor{SceneTint}
Bonsai & FP32 & 28.60 & 0.8555 & 0.3257 & 241.4 \\*
\rowcolor{SceneTint}
Bonsai & FP16 & 28.60 & 0.8555 & 0.3257 & 120.7 \\*
\rowcolor{SceneTint}
Bonsai & Atlas & 28.59 & 0.8550 & 0.3261 & 53.2 \\
\SceneSeparator
\rowcolor{white}
Train & FP32 & 20.94 & 0.7253 & 0.3403 & 241.4 \\*
\rowcolor{white}
Train & FP16 & 20.94 & 0.7253 & 0.3403 & 120.7 \\*
\rowcolor{white}
Train & Atlas & 20.93 & 0.7245 & 0.3407 & 53.2 \\
\SceneSeparator
\rowcolor{SceneTint}
Truck & FP32 & 24.13 & 0.8096 & 0.2620 & 362.2 \\*
\rowcolor{SceneTint}
Truck & FP16 & 24.13 & 0.8096 & 0.2620 & 181.1 \\*
\rowcolor{SceneTint}
Truck & Atlas & 24.12 & 0.8090 & 0.2624 & 79.6 \\
\SceneSeparator
\rowcolor{white}
Playroom & FP32 & 28.29 & 0.8785 & 0.3188 & 241.4 \\*
\rowcolor{white}
Playroom & FP16 & 28.29 & 0.8785 & 0.3188 & 120.7 \\*
\rowcolor{white}
Playroom & Atlas & 28.29 & 0.8781 & 0.3191 & 53.2 \\
\SceneSeparator
\rowcolor{SceneTint}
DrJohnson & FP32 & 28.17 & 0.8765 & 0.3137 & 362.2 \\*
\rowcolor{SceneTint}
DrJohnson & FP16 & 28.17 & 0.8765 & 0.3137 & 181.1 \\*
\rowcolor{SceneTint}
DrJohnson & Atlas & 28.16 & 0.8762 & 0.3141 & 79.6 \\
\bottomrule
\end{longtable}
\endgroup

%% file: sections/appendix_visual.tex
\clearpage
\section{Cross-Budget Visual Comparisons on Other Scenes}
\label{app:pareto-visual}
The following figures use the same method columns and budget rows as Figure~\ref{fig:pareto-qualitative}. Each panel labels its actual Gaussian count; two scenes appear on each page.

\begin{figure}[!ht]
\centering
\includegraphics[width=.96\linewidth]{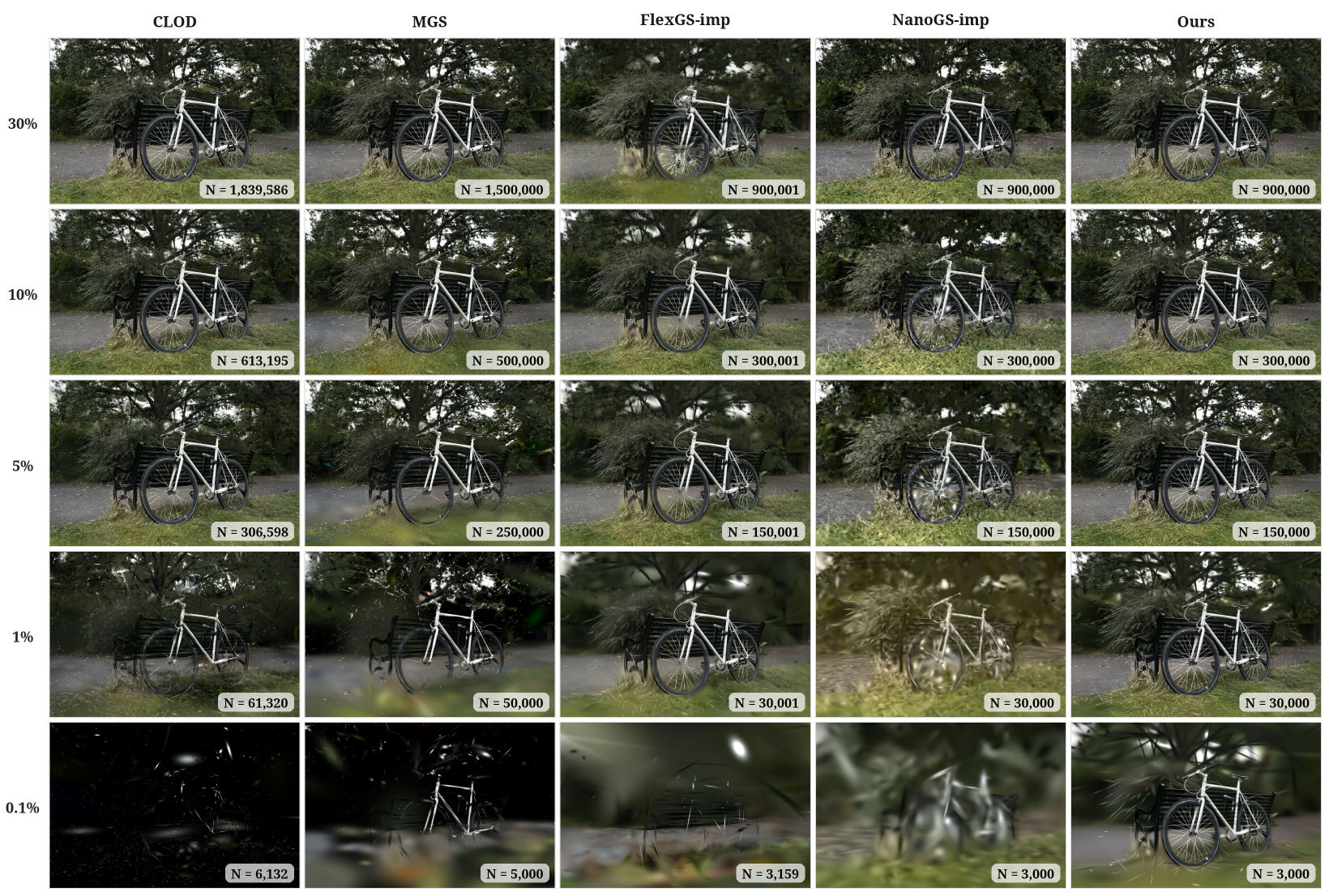}
\par\vspace{.5\baselineskip}
\includegraphics[width=.96\linewidth]{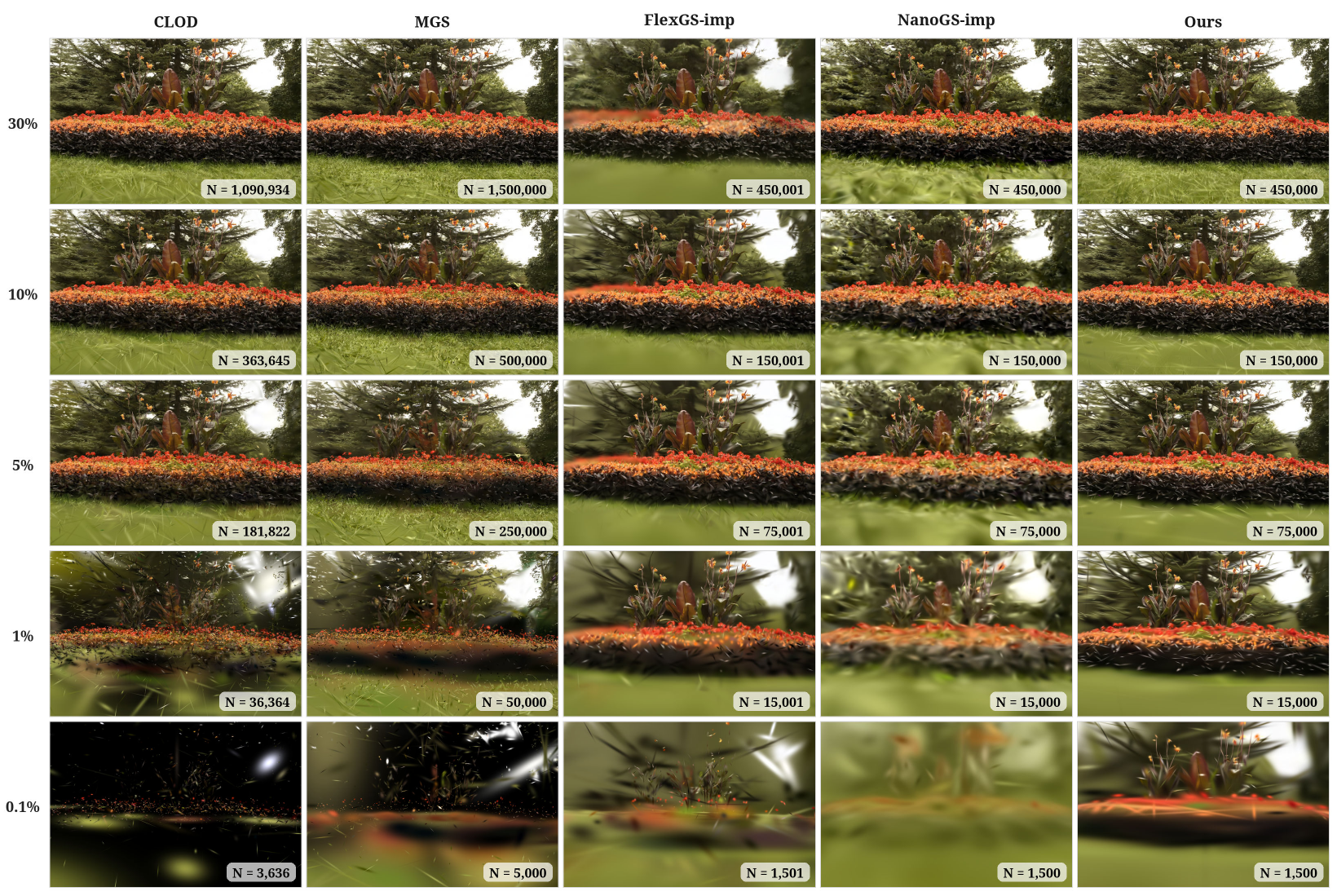}
\caption{Cross-budget reconstruction on Bicycle (top) and Flowers (bottom).}
\label{fig:pareto-extra-bicycle-flowers}
\end{figure}
\clearpage

\begin{figure}[!ht]
\centering
\includegraphics[width=.96\linewidth]{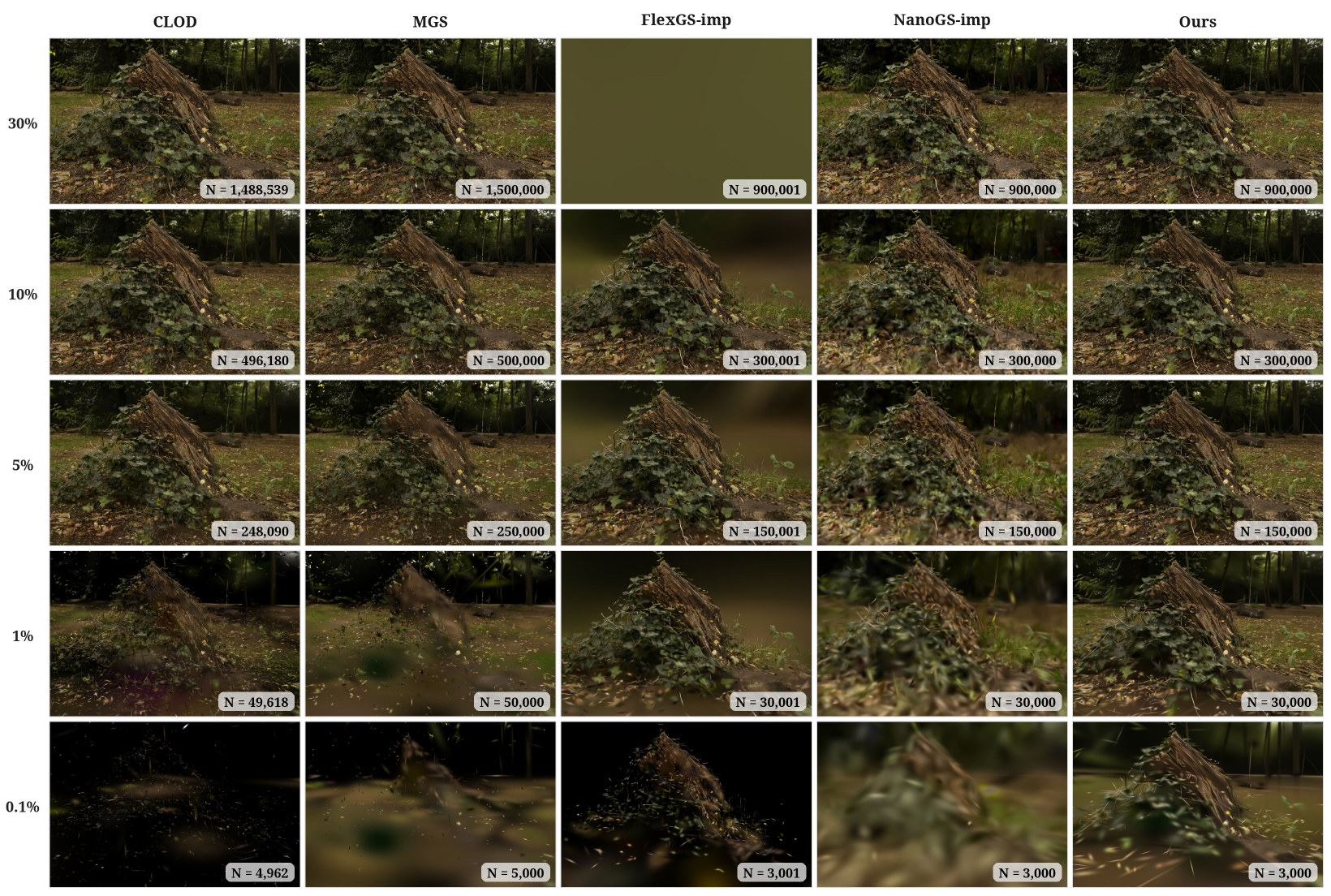}
\par\vspace{.5\baselineskip}
\includegraphics[width=.96\linewidth]{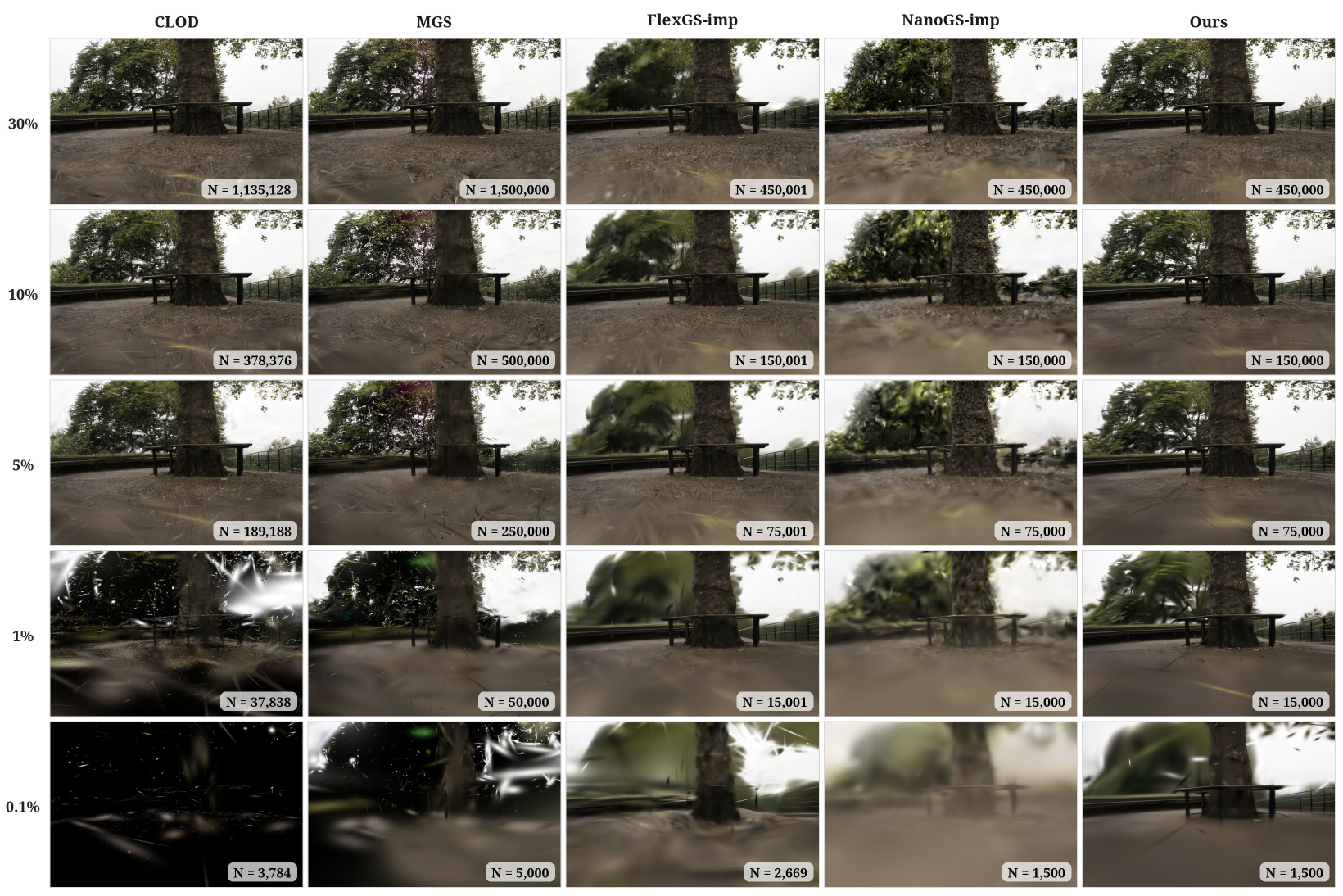}
\caption{Cross-budget reconstruction on Stump (top) and Treehill (bottom).}
\label{fig:pareto-extra-stump-treehill}
\end{figure}
\clearpage

\begin{figure}[!ht]
\centering
\includegraphics[width=.96\linewidth]{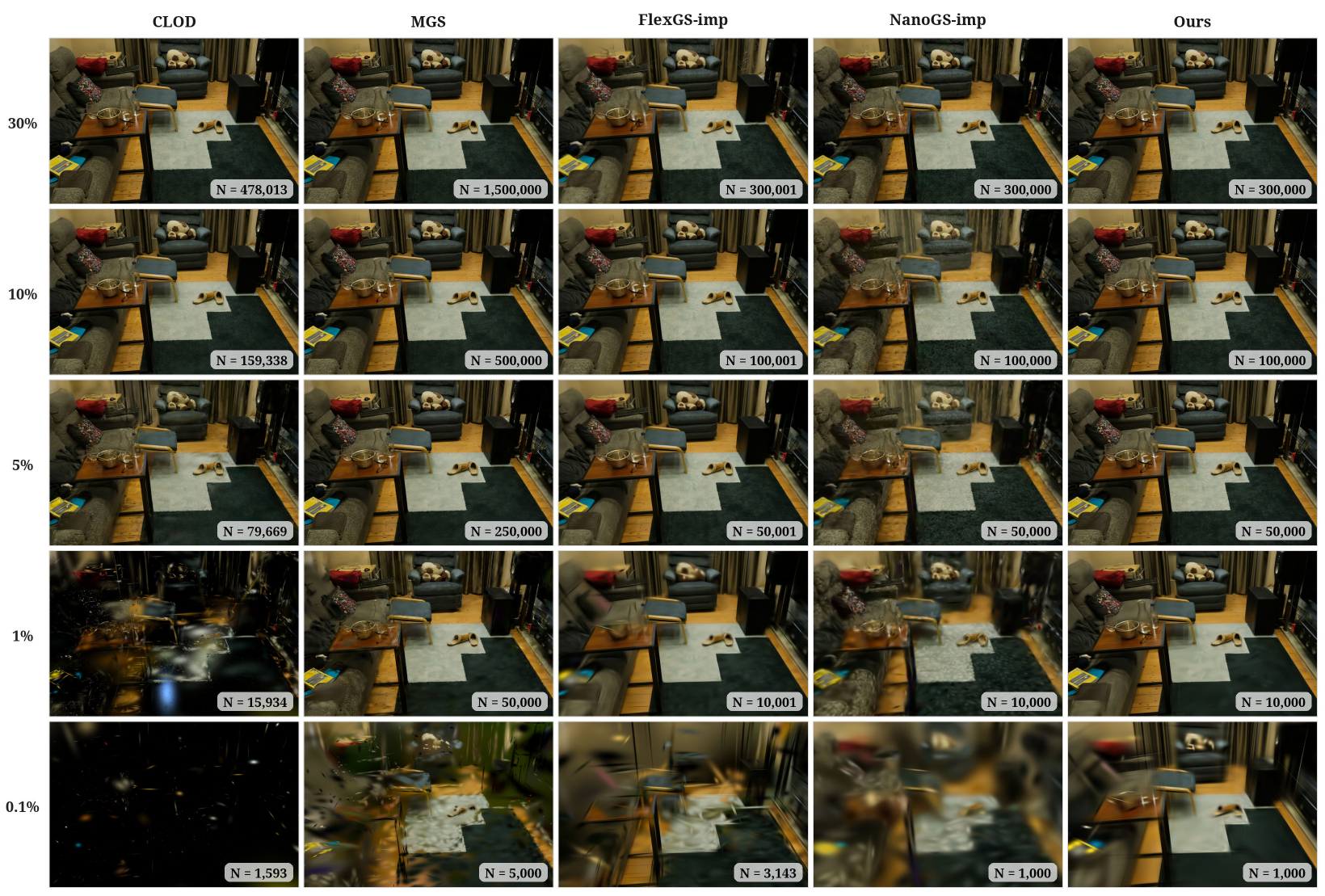}
\par\vspace{.5\baselineskip}
\includegraphics[width=.96\linewidth]{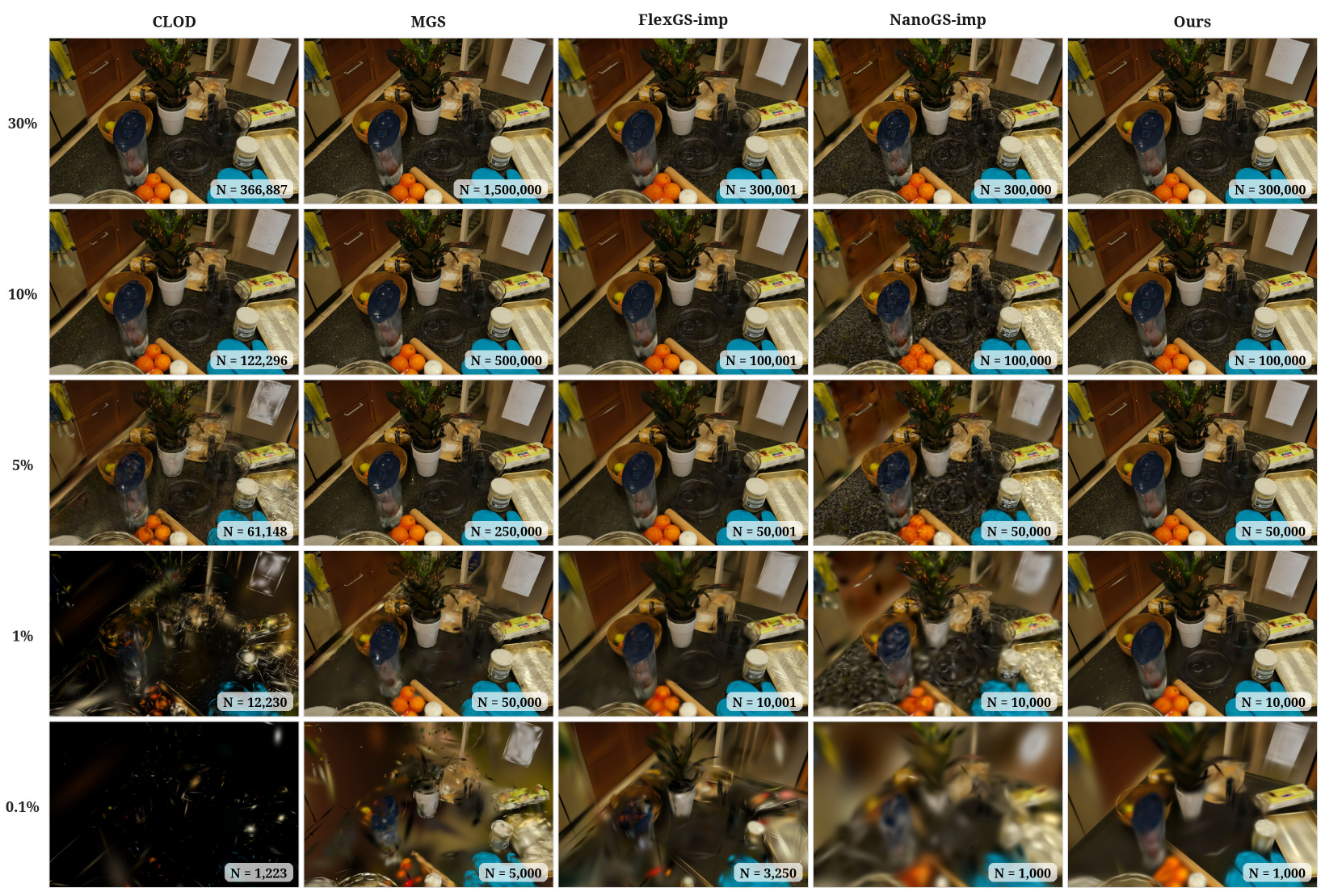}
\caption{Cross-budget reconstruction on Room (top) and Counter (bottom).}
\label{fig:pareto-extra-room-counter}
\end{figure}
\clearpage

\begin{figure}[!ht]
\centering
\includegraphics[width=.96\linewidth]{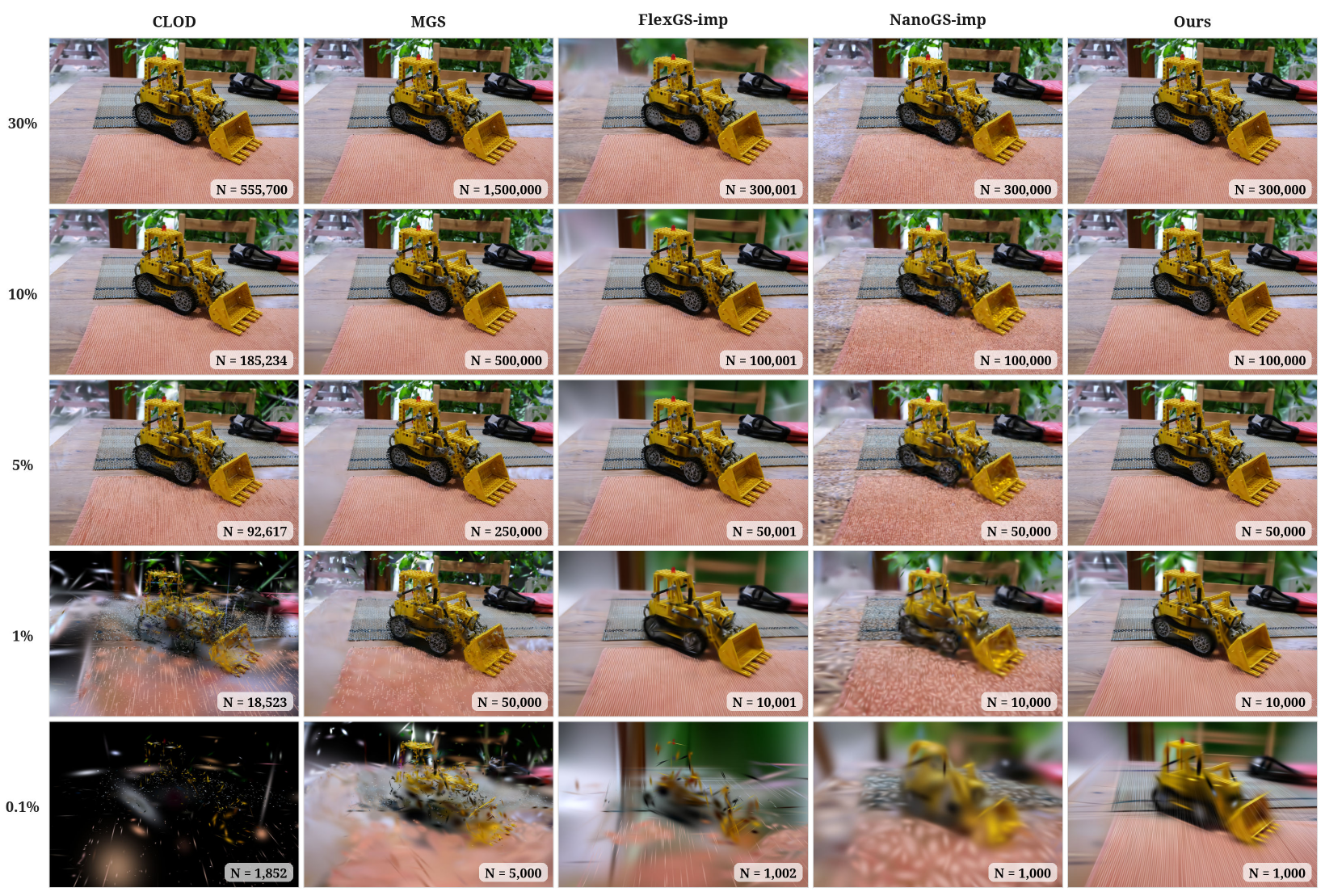}
\par\vspace{.5\baselineskip}
\includegraphics[width=.96\linewidth]{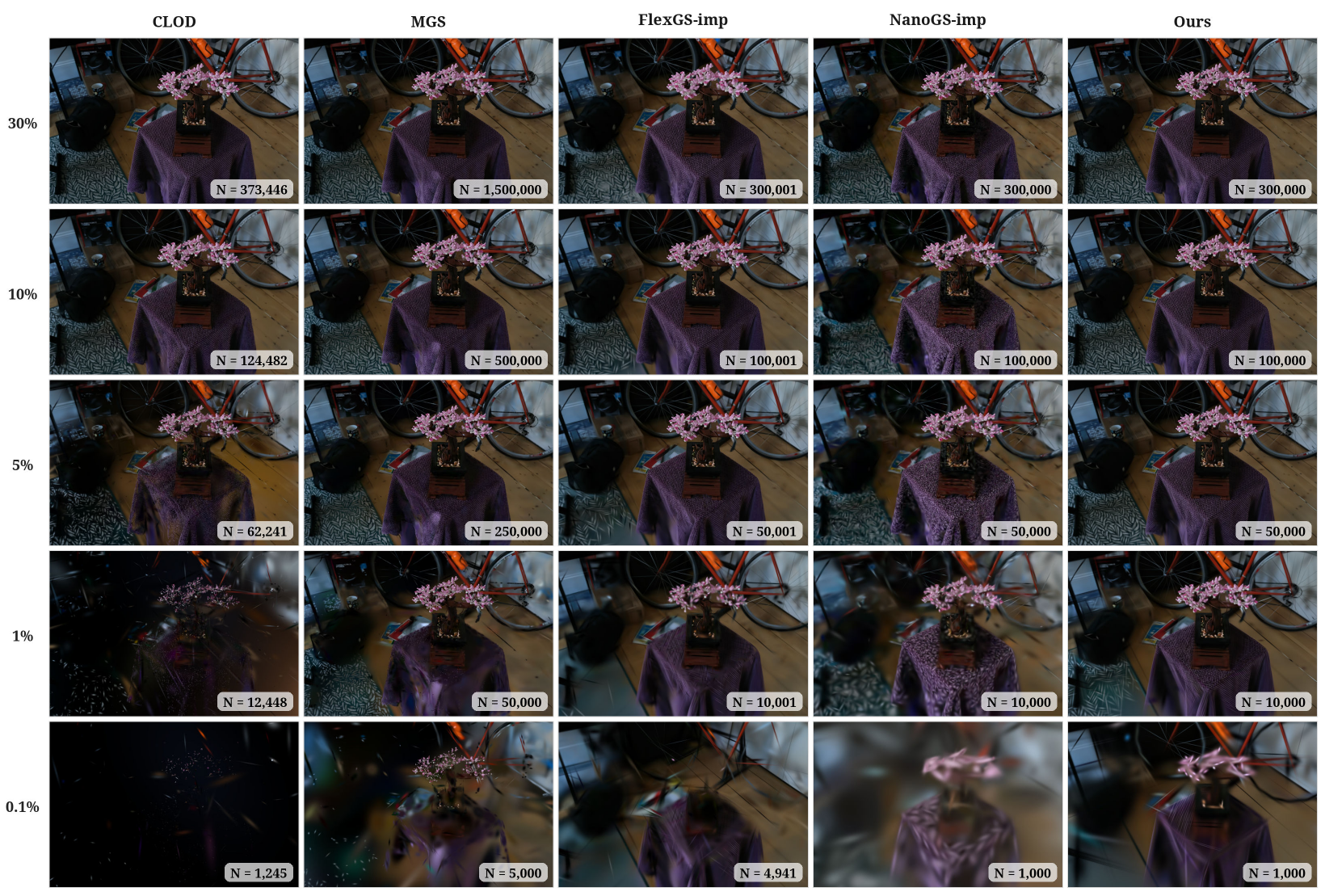}
\caption{Cross-budget reconstruction on Kitchen (top) and Bonsai (bottom).}
\label{fig:pareto-extra-kitchen-bonsai}
\end{figure}
\clearpage

\begin{figure}[!ht]
\centering
\includegraphics[width=.96\linewidth]{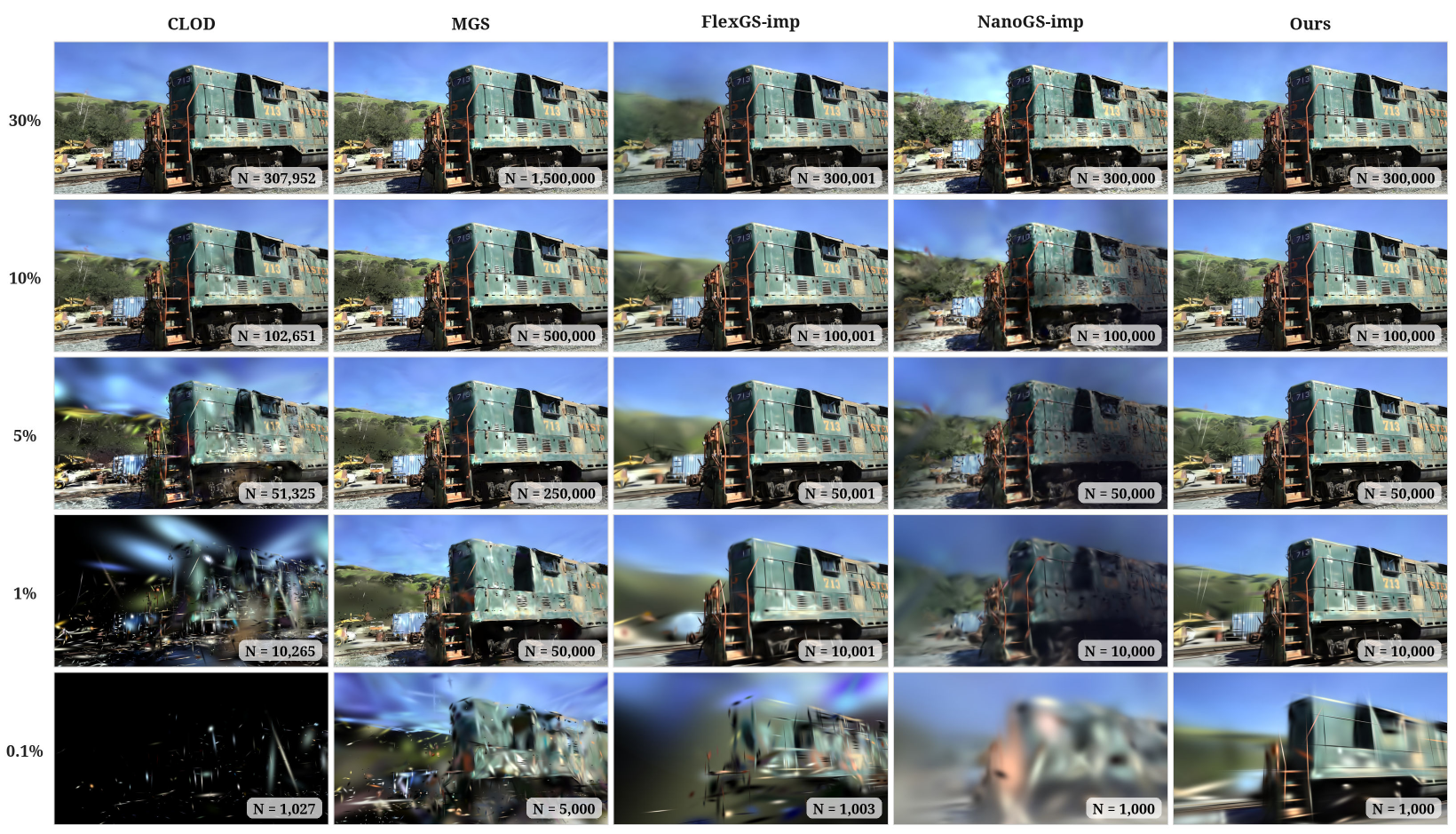}
\par\vspace{.5\baselineskip}
\includegraphics[width=.96\linewidth]{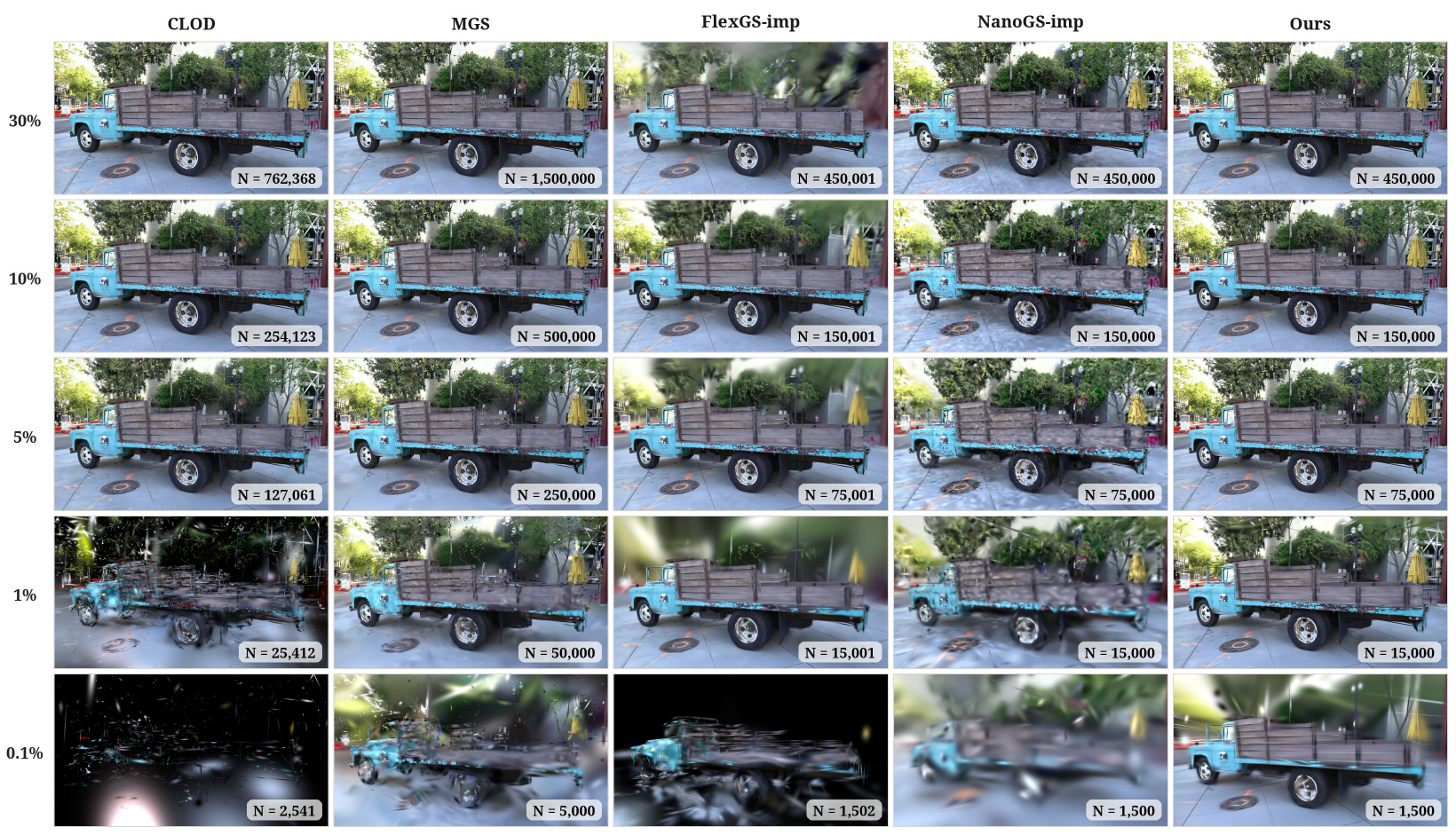}
\caption{Cross-budget reconstruction on Train (top) and Truck (bottom).}
\label{fig:pareto-extra-train-truck}
\end{figure}
\clearpage

\begin{figure}[!ht]
\centering
\includegraphics[width=.96\linewidth]{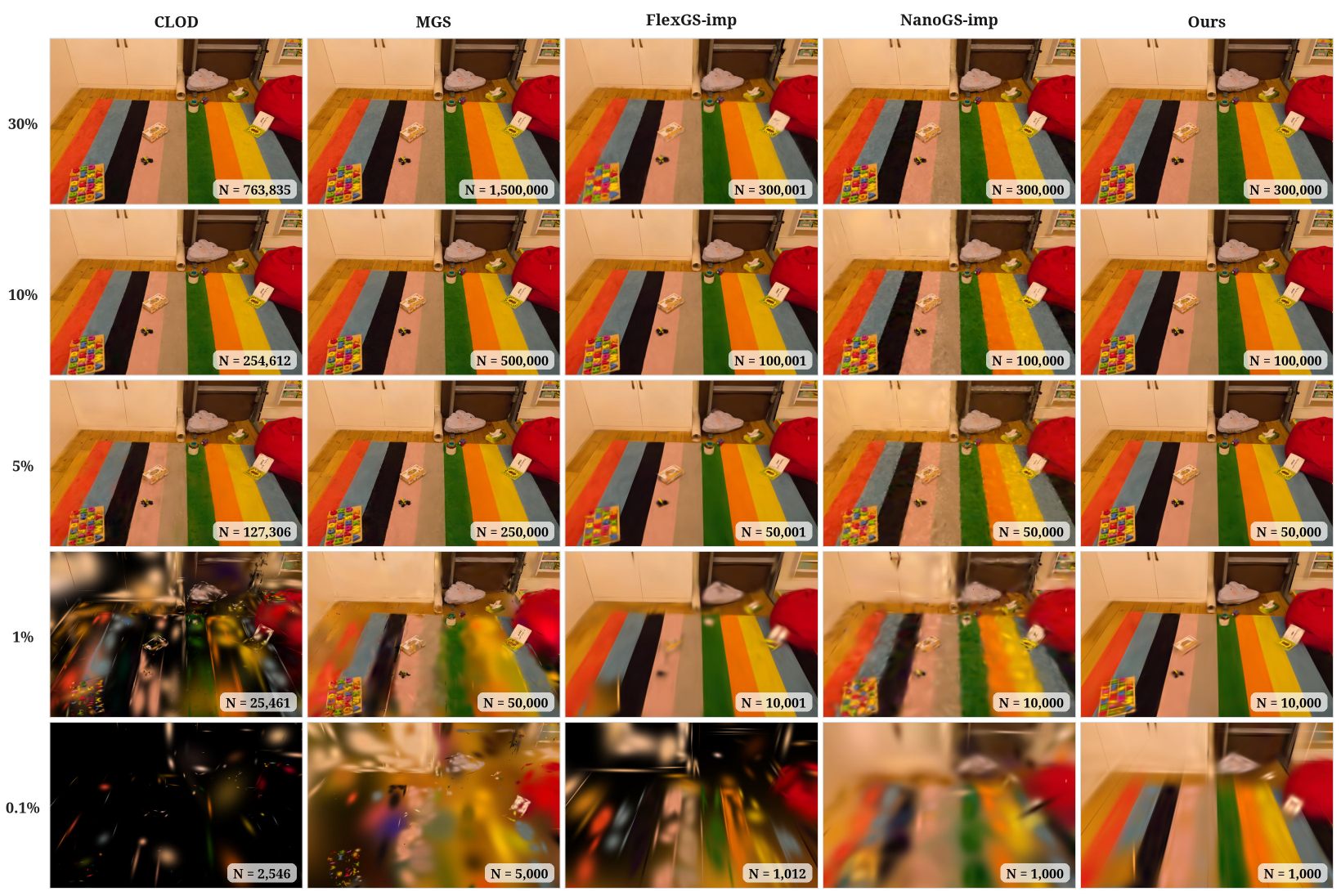}
\par\vspace{.5\baselineskip}
\includegraphics[width=.96\linewidth]{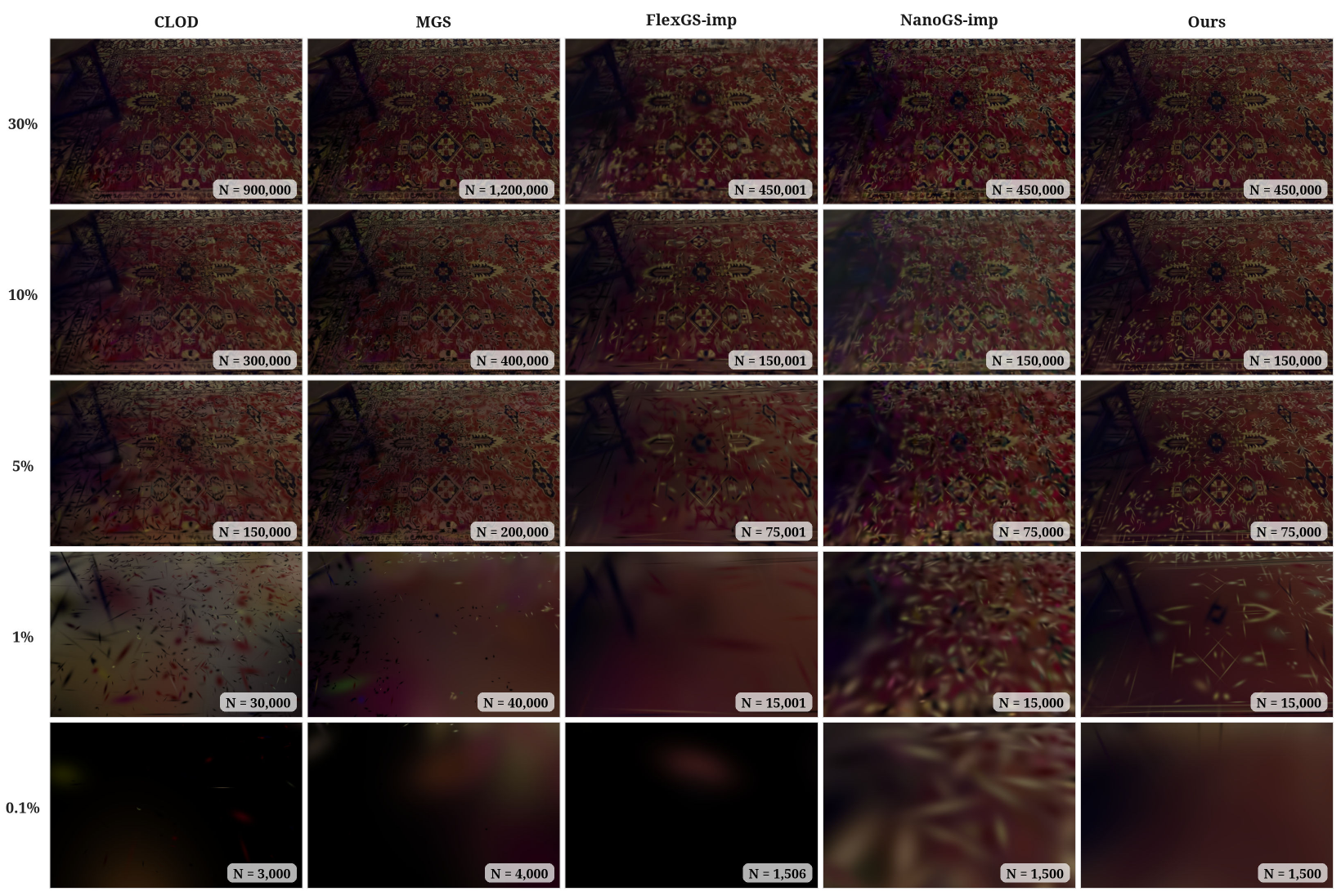}
\caption{Cross-budget reconstruction on Playroom (top) and DrJohnson (bottom).}
\label{fig:pareto-extra-playroom-drjohnson}
\end{figure}
\clearpage

%% file: sections/appendix_curves.tex
\section{Continuous-Budget Quality Curves on Other Scenes}
\label{app:pareto-curves}
\begin{figure}[!ht]
\centering
\noindent\makebox[.96\linewidth][l]{\textbf{Bicycle}}\par
\includegraphics[width=.96\linewidth]{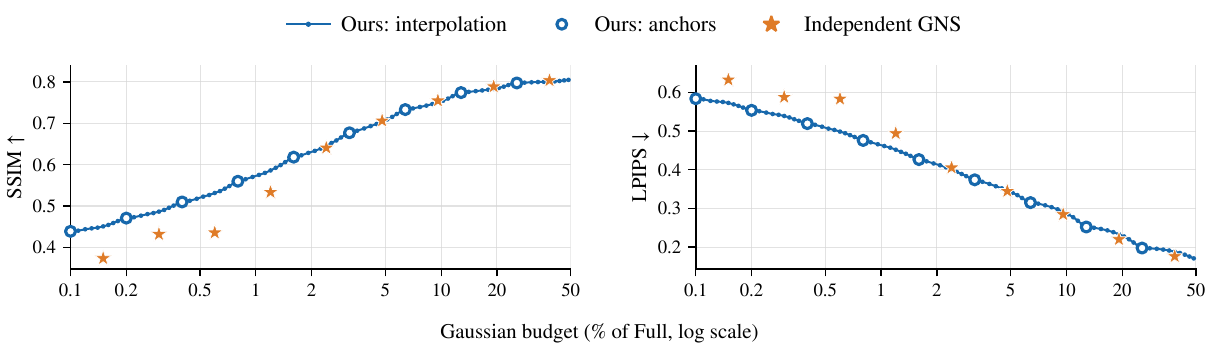}
\par\vspace{.25\baselineskip}
\noindent\makebox[.96\linewidth][l]{\textbf{Flowers}}\par
\includegraphics[width=.96\linewidth]{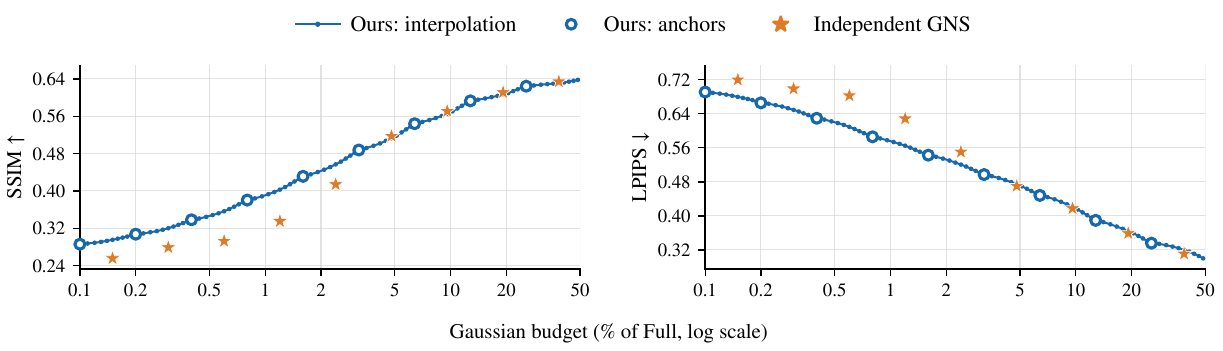}
\par\vspace{.25\baselineskip}
\noindent\makebox[.96\linewidth][l]{\textbf{Stump}}\par
\includegraphics[width=.96\linewidth]{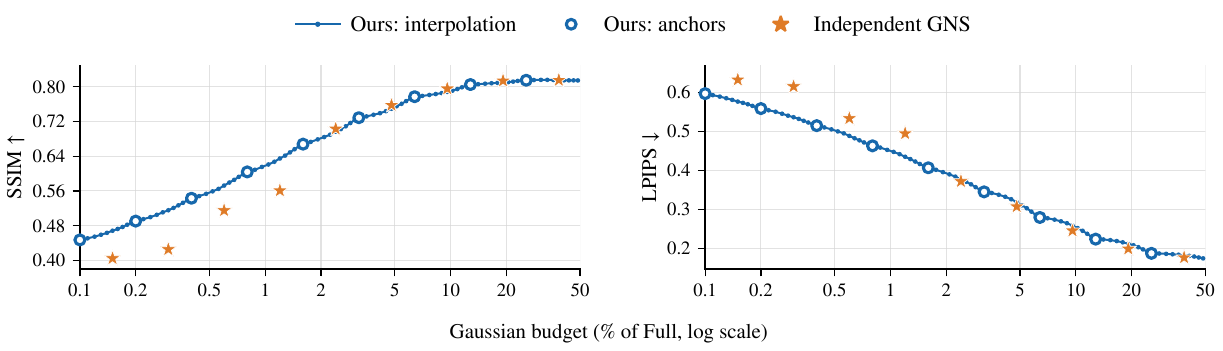}
\par\vspace{.25\baselineskip}
\noindent\makebox[.96\linewidth][l]{\textbf{Treehill}}\par
\includegraphics[width=.96\linewidth]{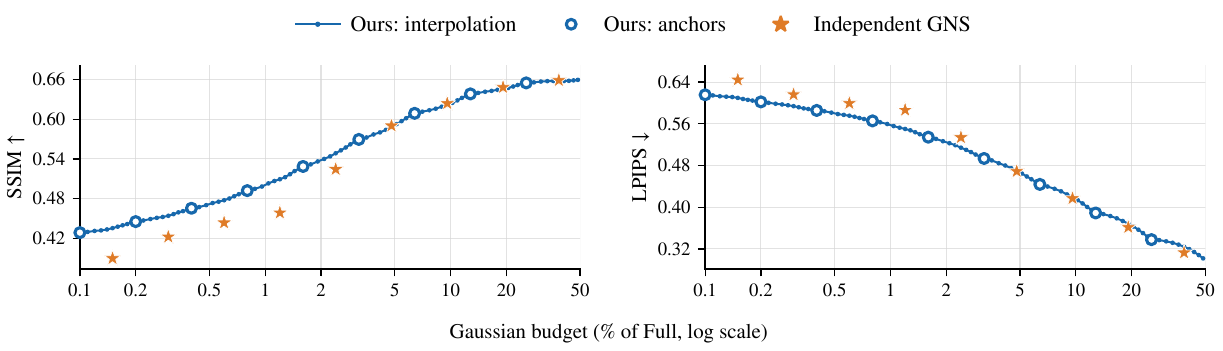}
\caption{Continuous-budget quality curves for Bicycle, Flowers, Stump, and Treehill, top to bottom.}
\label{fig:pareto-curves-outdoor}
\end{figure}

\begin{figure}[!ht]
\centering
\noindent\makebox[.96\linewidth][l]{\textbf{Room}}\par
\includegraphics[width=.96\linewidth]{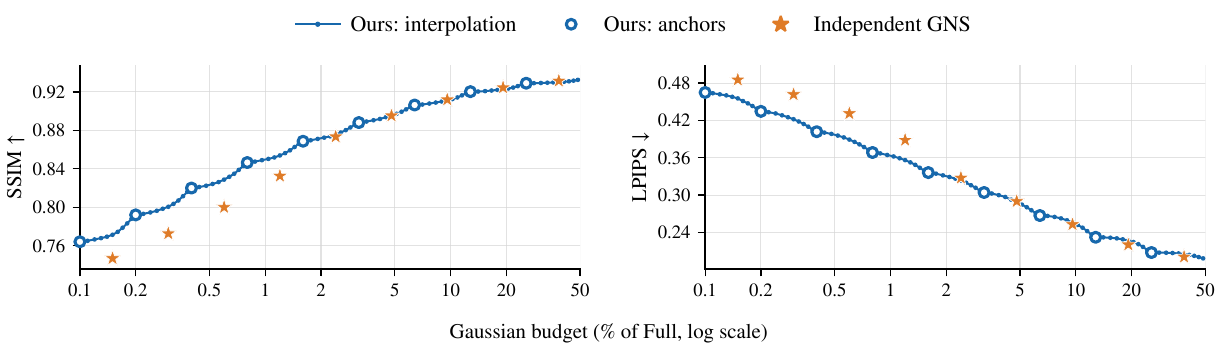}
\par\vspace{.25\baselineskip}
\noindent\makebox[.96\linewidth][l]{\textbf{Counter}}\par
\includegraphics[width=.96\linewidth]{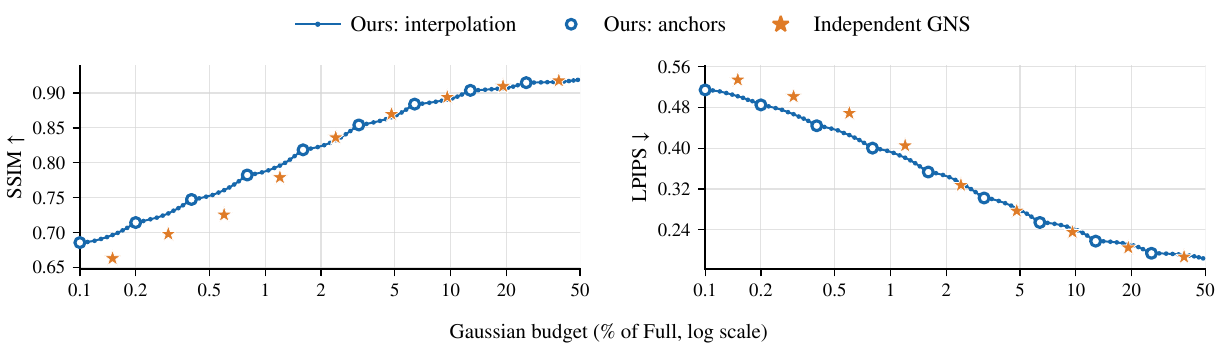}
\par\vspace{.25\baselineskip}
\noindent\makebox[.96\linewidth][l]{\textbf{Kitchen}}\par
\includegraphics[width=.96\linewidth]{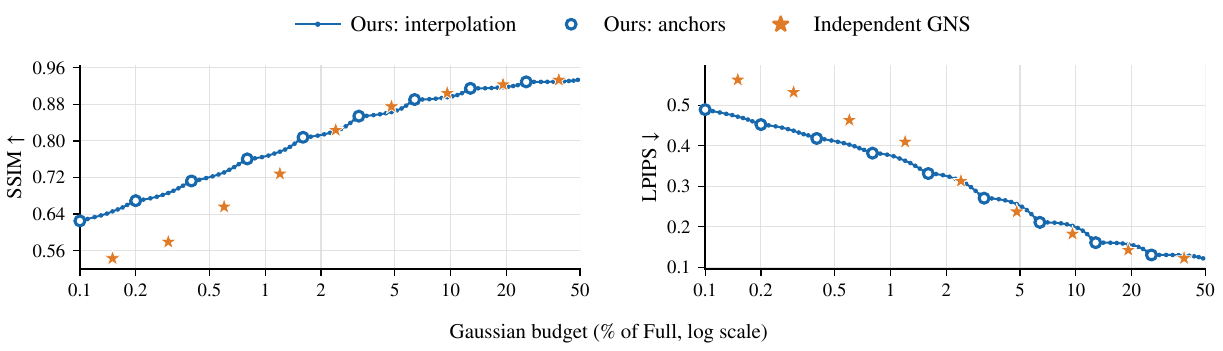}
\par\vspace{.25\baselineskip}
\noindent\makebox[.96\linewidth][l]{\textbf{Bonsai}}\par
\includegraphics[width=.96\linewidth]{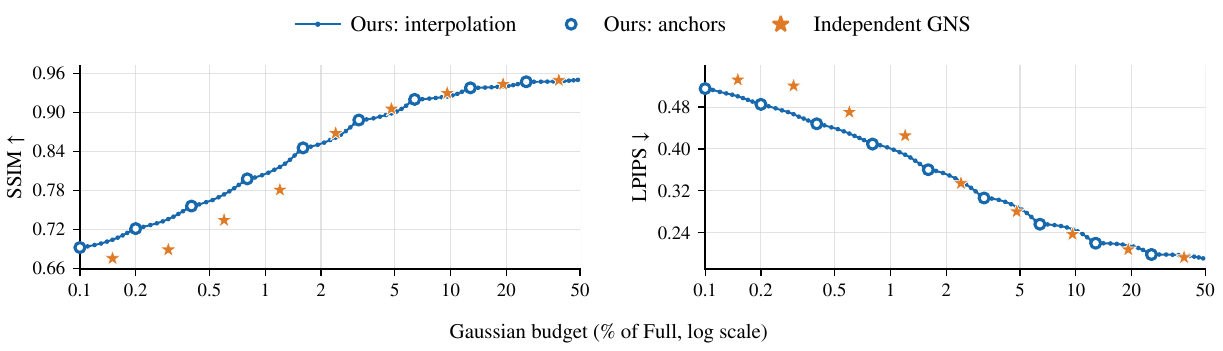}
\caption{Continuous-budget quality curves for Room, Counter, Kitchen, and Bonsai, top to bottom.}
\label{fig:pareto-curves-indoor}
\end{figure}

\begin{figure}[!ht]
\centering
\noindent\makebox[.96\linewidth][l]{\textbf{Train}}\par
\includegraphics[width=.96\linewidth]{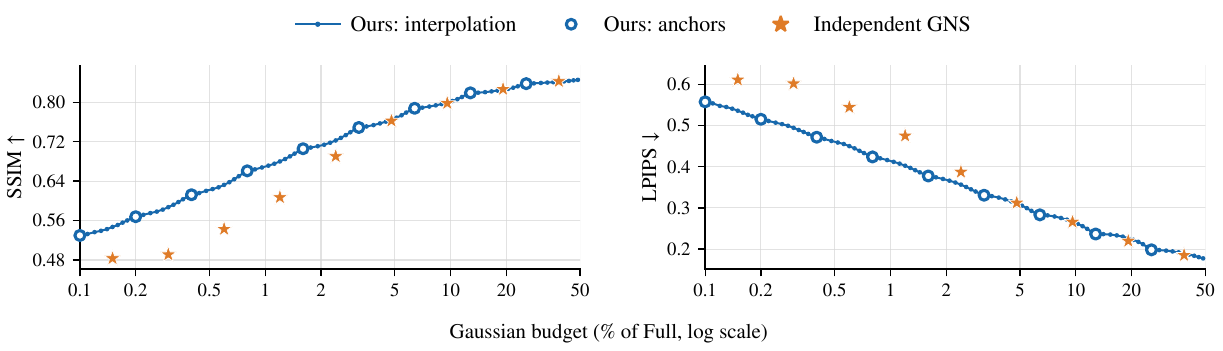}
\par\vspace{.25\baselineskip}
\noindent\makebox[.96\linewidth][l]{\textbf{Truck}}\par
\includegraphics[width=.96\linewidth]{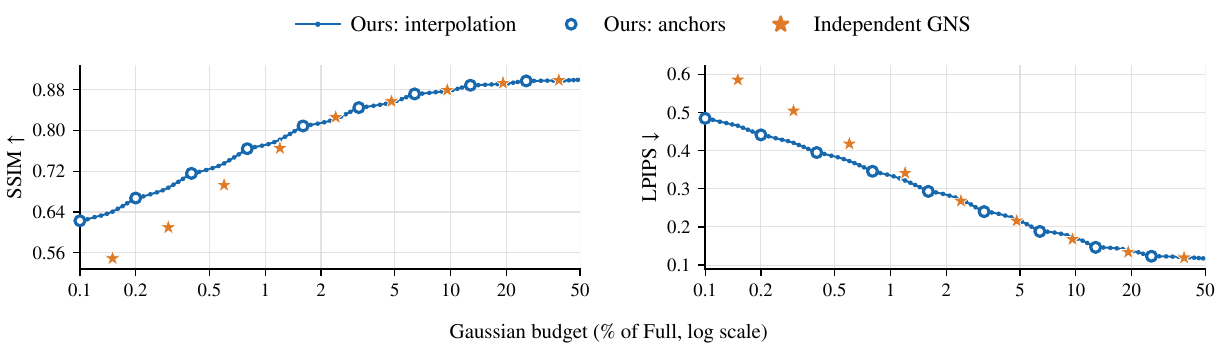}
\par\vspace{.25\baselineskip}
\noindent\makebox[.96\linewidth][l]{\textbf{Playroom}}\par
\includegraphics[width=.96\linewidth]{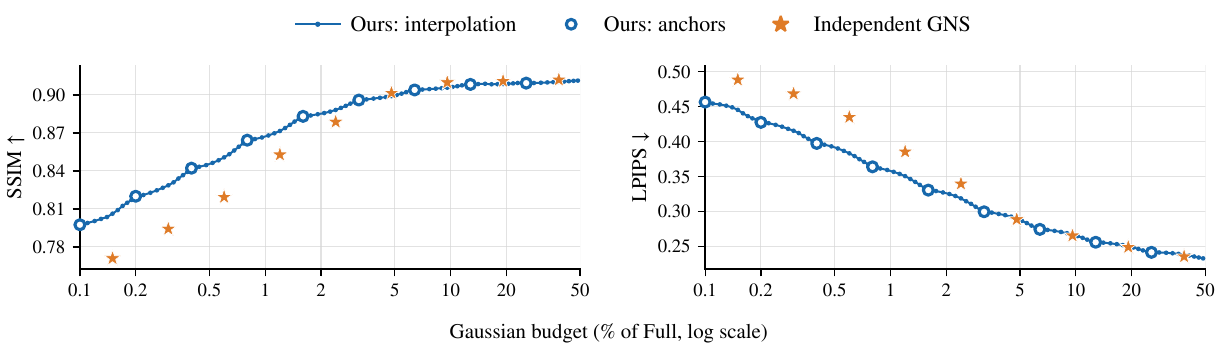}
\par\vspace{.25\baselineskip}
\noindent\makebox[.96\linewidth][l]{\textbf{DrJohnson}}\par
\includegraphics[width=.96\linewidth]{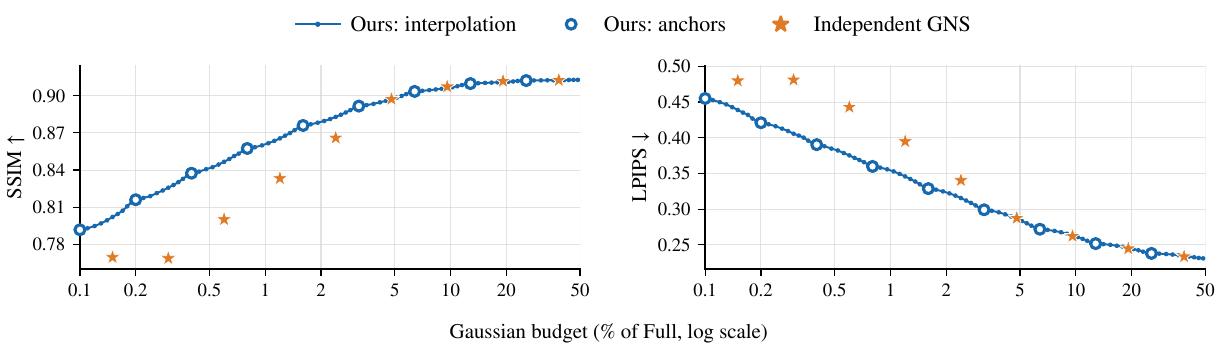}
\caption{Continuous-budget quality curves for Train, Truck, Playroom, and DrJohnson, top to bottom.}
\label{fig:pareto-curves-mixed}
\end{figure}

Figures~\ref{fig:pareto-curves-outdoor}--\ref{fig:pareto-curves-mixed} show anchors, actual interval-decoding results, and independent GNS references for the 12 scenes beyond Garden in Figure~\ref{fig:pareto-continuous}.

\clearpage

\clearpage

\clearpage